\documentclass[12pt]{elsarticle}

\usepackage{graphicx}
\usepackage{amssymb}

\usepackage{lineno}
\usepackage{color}
\usepackage{courier}
\usepackage{xcolor}
\usepackage{tikz}
\usetikzlibrary{calc}
\usetikzlibrary{backgrounds,positioning}
\usetikzlibrary{decorations.pathreplacing}
\usepackage{diagbox}
\usepackage{caption}
\usepackage{rotating} 
\usepackage{subcaption}
\usepackage{xcolor}
\usepackage[linesnumbered,ruled,vlined,procnumbered]{algorithm2e}
\usepackage{etoolbox}

\makeatletter
\AtBeginEnvironment{procedure}{\let\c@algocf\c@procedure}
\makeatother
\usepackage{amsmath}
\DeclareMathOperator*{\argmax}{argmax}
\DeclareMathOperator*{\argmin}{argmin}

\SetCommentSty{mycommfont}

\SetKwInput{KwInput}{Input}                
\SetKwInput{KwOutput}{Output}

\tikzset{
	rednode/.style = {rectangle, draw=black!60, fill=red!20, very thick, minimum size=5mm,outer sep=0pt},
	bluenode/.style = {rectangle, draw=black!60, fill=blue!20, very thick, minimum size=5mm,outer sep=0pt},
	greennode/.style = {rectangle, draw=black!60, fill=green!20, very thick, minimum size=5mm,outer sep=0pt},
	whitenode/.style = {rectangle, draw=black!60, fill=white!20, very thick, minimum size=5mm,outer sep=0pt},
}

\makeatletter
\def\ps@pprintTitle{%
 \let\@oddhead\@empty
 \let\@evenhead\@empty
 \def\@oddfoot{\centerline{\thepage}}%
 \let\@evenfoot\@oddfoot}
\makeatother

\begin{document}

\begin{frontmatter}


\title{On Population-Based Algorithms for Distributed Constraint Optimization Problems}


\author[du]{Saaduddin Mahmud}
\author[du]{Md. Mosaddek Khan}
\author[ip]{Nicholas R. Jennings}
\address[du]{Department of Computer Science and Engineering, University of Dhaka}
\address[ip]{Departments of Computing and Electrical and Electronic Engineering, Imperial College London}

\begin{abstract}
  \small{Distributed Constraint Optimization Problems (DCOPs) are a widely studied class of optimization problems in which interaction between a set of cooperative agents are modeled as a set of constraints. DCOPs are NP-hard and significant effort has been devoted to developing methods for finding incomplete solutions. In this paper, we study an emerging class of such incomplete algorithms that are broadly termed as population-based algorithms. The main characteristic of these algorithms is that they maintain a population of candidate solutions of a given problem and use this population to cover a large area of the search space and to avoid local-optima. In recent years, this class of algorithms has gained significant attention due to their ability to produce high-quality incomplete solutions. With the primary goal of further improving the quality of solutions compared to the state-of-the-art incomplete DCOP algorithms, we present two new population-based algorithms in this paper. Our first approach, Anytime Evolutionary DCOP or AED, exploits evolutionary optimization meta-heuristics to solve DCOPs. We also present a novel anytime update mechanism that gives AED its anytime property. While in our second contribution, we show that population-based approaches can be combined with local search approaches. Specifically, we develop an algorithm called DPSA based on the Simulated Annealing meta-heuristic. We empirically evaluate these two algorithms to illustrate their respective effectiveness in different settings against the state-of-the-art incomplete DCOP algorithms including all existing population-based algorithms in a wide variety of benchmarks. Our evaluation shows AED and DPSA markedly outperform the state-of-the-art and produce up to 75\% improved\:solutions}.
\end{abstract}

\end{frontmatter}



\section{Introduction}
\noindent Distributed Constraint Optimization Problems (DCOPs) are a widely used framework for coordinating interactions in cooperative multi-agent systems (MAS). In particular, agents in this framework need to coordinate value assignments to their variables in such a way that minimizes constraint violations by optimizing their aggregated costs \cite{yokoo1998distributed}. DCOPs have gained popularity due to their applications in various real-world multi-agent coordination problems, including distributed meeting scheduling \cite{realworld}, coordinating mesh networks for maximizing signal strength \cite{RFBnch}, target tracking using mobile sensor agents \cite{farinelli2014agent,MST}\:and economic dispatch and demand response in smart grids \cite{fioretto2017distributed}.\par

Over the last two decades, several algorithms have been proposed to solve DCOPs, and they can be broadly classified into two classes: complete and incomplete. The former always provide an optimal solution of a given DCOP. These algorithms can be further classified as search-based \cite{Hirayama1997DistributedPC, modi2005adopt, AMIRGER2010AsynchronousFB,netzer2012concurrent,NCBB,optapo,Yeoh2008BnBADOPTAA} and inference-based \cite{Petcu2005ASM,Petcu2007MBDPOPAN,Petcu2006ODPOPAA,Petcu2005ASM,Vinyals2009GeneralizingDA,rashik2019speeding,RMBDPOP}. Since solving DCOPs optimally is NP-hard \cite{modi2005adopt}, scalability becomes an issue as the system grows for the complete algorithms. In contrast, incomplete algorithms compromise some solution quality for scalability. As a consequence, diverse classes of incomplete algorithms have been developed to deal with large-scale DCOPs. Until recently, these incomplete algorithms could be classified into three classes: local search-based algorithms \cite{zhang2005distributed, maheswaran2004distributed, okamoto2016distributed, ArshadDistributedSA}, inference-based algorithms \cite{farinelli2008decentralised, speed, zivan2012max, Rogers2011BoundedAD,Cohen2017MaxsumRT} and sample-based algorithms \cite{Ottens2012DUCTAU, dgibbs}.\par

In this paper, we study a new class of incomplete DCOP algorithms that have recently emerged in the literature under the general heading of population-based algorithms. This class of algorithms keep a set of candidate solutions of the DCOP problem being solved and exploit them to cover a large search space and avoid local-optima. This is in contrast to the previous three classes of incomplete algorithm where only a single candidate solution is updated by all the agents. This, in turn, results in focusing on a narrower search space which adversely affects exploration and can cause premature convergence to poor local-optima.\par

The population-based approaches started with ACO\_DCOP \cite{Chen2018AnAA} which adapts a centralized population-based approach using as Ant Colony Optimization (ACO) procedure \cite{Dorigo2006AntCO}. It is worth noting that, in addition to ACO, a wide variety of other centralized population-based algorithms exist and a large number of them go under the broad heading of evolutionary optimization techniques \cite{holland1992adaptation,ep}. These techniques have proven very effective in solving NP-hard combinatorial optimization problems \cite{Fogel1988AnEA, tsang1990applying, Hirayama1997DistributedPC}. However, no prior work has adapted evolutionary optimization techniques to solve DCOPs. The effectiveness of these techniques in solving combinatorial optimization problems, along with the potential of the population-based DCOP solver, motivate our work in this nascent area.\par 

In more detail, we propose a new population-based algorithm that uses evolutionary optimization to solve DCOPs. We call this Anytime Evolutionary DCOP (AED). AED maintains a set of candidate solutions that are distributed among the agents, and they iteratively search for new improved solutions by locally modifying the candidate solutions (reproduction). During this iterative process, only the most promising subset of the candidate solutions are kept, and the rest are discarded through a stochastic selection process. This helps AED to focus toward more promising areas of the search space. Moreover, we introduce a novel anytime update mechanism to identify the best among this distributed set of candidate solutions and help the agents to coordinate value assignments to their variables based on the best candidate solution. Our theoretical analysis proves that AED is anytime. Our experimental results show that AED can produce a solution of better quality compared to the aforementioned algorithms. Further, the axiomatic nature of AED enables it to effectively deal with DCOPs having global constraints (e.g. All Different \cite{van2001alldifferent}), and our empirical evidence illustrates AED's efficacy in doing so. 

Even though population-based algorithms find high quality solution compared to other incomplete algorithms, they generally incur large computation and communication costs compared to the algorithms that only maintain a single candidate solution. Therefore, scalability, although significantly better than complete algorithms, can become an issue for large problem instances. On the other hand, existing local search algorithms are highly scalable; however, the solution quality produced by these algorithms can be poor compared to population-based algorithms (see Section 6 for results). In the wake of this trade-off, in our second contribution, we focus on the hybridization between population-based algorithm and local search algorithms to get the benefit of both paradigms. Although, with the use of local search approach, unlike AED, it is no longer able to deal with DCOPs with global constraints.\par 
 
 In more detail, we propose a novel hybrid algorithm that we call Distributed Parallel Simulated Annealing (DPSA). DPSA is based on Simulated Annealing (SA), a meta-heuristic motivated by a physical analogy of controlled temperature cooling (i.e. annealing) of a material \cite{Kirkpatrick1983OptimizationBS}. DPSA maintains a set of candidate solutions which are modified independently by applying a local search approach in parallel. However, information gathered from these independent runs of the local search is exploited to improve the configuration of the search parameter. This helps DPSA avoid bad local-optima and find a good balance between exploration and exploitation. Further, parallelization helps DPSA cover a large portion of the search space. This results in a significant improvement of solution quality compared to the state-of-the-art DCOPs algorithms, while being significantly more scalable.\par 

To summarize our contribution in this paper:
\begin{itemize}
    \item We present a novel population-based algorithm called AED that adapts evolutionary optimization techniques for solving DCOPs.    
    \item We also propose a novel hybrid population-based approach called DPSA that improves the scalability of population-based algorithms while producing a solution of state-of-the-art quality. DPSA uses a Monte Carlo importance sampling method called Cross-Entropy (CE) sampling \cite{Kroese2011HandbookOM} to learn the optimal parameter configuration. However, much of its success depends on prior knowledge about the parameter search space which may not always be available (depending on the deployed application). To address this issue, we present a variant of DPSA that can be used on those applications where no prior knowledge is available. We call it DPSA Greedy Baseline (DPSA\_GB).
    \item Finally, we extensively evaluate these algorithms with the current state-of-the-art incomplete algorithms in a wide range of benchmarks. Our evaluation shows both of our proposed algorithms produce anytime performance that outperforms the state-of-the-art incomplete DCOP algorithms. Our empirical results also illustrate the respective effectiveness of our proposed algorithms in different settings. 
\end{itemize}
The rest of the paper is organized as follows. We discuss existing related works on incomplete DCOPs in Section 2. The section that follows provides the necessary background that gives context to our proposed algorithms. In Sections 4 and 5, we describe AED and DPSA in detail, respectively. Section 6 presents a wide variety of experiments that evaluates AED and DPSA against the state-of-the-art. Finally, we conclude with Section 7 where we summarize our findings and discuss future work.\par

\section{Related Work}
\noindent Over the last two decades, a number of complete algorithms have been proposed for solving DCOPs. Among the complete algorithms SyncBB \cite{Hirayama1997DistributedPC}, NCBB \cite{NCBB}, AFB \cite{AMIRGER2010AsynchronousFB}, ConFB\cite{netzer2012concurrent}, ADOPT \cite{modi2005adopt}, BnB-ADOPT \cite{Yeoh2008BnBADOPTAA}, OptAPO \cite{optapo}, DPOP \cite{Petcu2005ASM} , ODPOP \cite{Petcu2006ODPOPAA}, MB-DPOP \cite{Petcu2007MBDPOPAN}, RMB-DPOP \cite{RMBDPOP}, Action-GDL \cite{AGDL}, PT-FB \cite{litov2017forward} and HS-CAI \cite{Chen2020HSCAIAH} are the most well known. A considerable amount of effort has been made to improve the scalability of complete DCOPs algorithms. However, solving DCOPs optimally is NP-Hard. In practice, this means that the use of these algorithms are largely restricted to small instances of DCOPs (i.e. small number of agents and constraints). This motivated a separate line of research work that studies incomplete DCOP algorithms which is the primary focus of this paper. In this section, we are going to discuss the existing body of work on incomplete DCOPs algorithms in more detail.\par

As mentioned above, until recently, incomplete algorithms were separated into three different categories: (i) local search-based, (ii) inference-based and (iii) sampling-based. Local search-based algorithms are the most common approach. The most well-known and simplest algorithm of this class is DSA \cite{fitzpatrick2003distributed}. DSA is an iterative algorithm where each agent makes the best assignment to the variables they control based on the current assignment of its neighbours with a static probability. DSAN \cite{ArshadDistributedSA} improves upon DSA by incorporating simulated annealing to calculate this probability rather than using a static probability. DSAN is of particular interest to our work since it is the first algorithm to consider Simulated Annealing for solving DCOPs and can be considered a direct predecessor of DPSA. The main difference is that DPSA keeps a population that runs DSAN in parallel and exploits the population to find the best parameter configuration to solve a given DCOP instance. Other notable examples of local search-based algorithms are MGM and MGM2 \cite{maheswaran2004distributed}, GDBA \cite{okamoto2016distributed} and DSA-SDP\cite{zivan2014explorative}. Also, to further enhance the solution quality and incorporate an anytime property, the Anytime Local Search (ALS) framework \cite{zivan2014explorative} has been introduced.\par

Inference-based algorithms, such as Max-Sum \cite{farinelli2008decentralised} or Fast Max-Sum \cite{Ramchurn2010}, operate on a factor graph representation of DCOPs. Max-Sum is particularly prominent as it can handle n-ary constraints explicitly and can guarantee convergence to an optimal solution in acyclic factor graphs. However, solutions produced by Max-Sum on cyclic factor graphs can be of poor quality. To address this issue, different variants were proposed.  Notably, Bounded Max-Sum \cite{Rogers2011BoundedAD} and Max-Sum\_ADVP \cite{zivan2012max} overcame these issues to a certain extent and produce better solutions than Max-Sum.\par

The third approach for finding a non-exact solution is based on sampling. In particular, \cite{Ottens2012DUCTAU} propose a distributed version (i.e. DUCT) of the popular UCT algorithm \cite{Kocsis2006BanditBM}. However, the memory requirement of DUCT grows exponentially with the number of agents, and that limits its applicability in large settings. To deal with this shortcoming, a sampling-based algorithm called D-Gibbs\cite{dgibbs} was proposed. D-Gibbs maps DCOP to MAP (Maximum a posterior) and uses a distributed variant of the well known Gibbs sampling \cite{geman1984stochastic}. D-Gibbs only requires a linear amount of memory while producing a solution of equal or better quality compared to DUCT.\par

Population-based algorithms recently emerged as a new category of incomplete algorithms through the introduction of ACO\_DCOP \cite{Chen2018AnAA}. Although there are prior examples of population-based DCSP solvers such as ACO\_DCSP \cite{ACODCSP} and SoHC \cite{dozier2007distributed}, they are not directly suitable for solving DCOPs. This is because solving DCOPs requires enumerating all possible assignment combinations to find the optimal solution rather than just finding a feasible solution. Population-based algorithms primarily gained attention for solving DCOPs because of their potential to produce incomplete solutions of higher quality compared to the previous three classes of algorithms \cite{Chen2018AnAA}. It is worth noting that there is a concurrent development (more specifically with \cite{AED, DPSACE}) in the literature named LSGA \cite{chen2020genetic}. LSGA runs local search algorithms in parallel and tries to combine multiple candidate solutions using genetic operations to improve exploration. However, as our experiments show, DPSA achieves a better balance between exploration and exploitation by learning meta-information (i.e. global parameter configuration) from parallel runs of the local search algorithms. 

\section{Background}
\noindent In this section, we first give a formal definition of DCOPs. In the subsequent three subsections, we provide a brief description of Distributed Simulated Annealing, Anytime Local Search and Cross-Entropy Sampling respectively which we use in our proposed algorithms, which we discuss in the sections that follow.
\subsection{Distributed Constraint Optimization Problems}
\noindent Formally, a DCOP is defined by a tuple $ \langle X,D,F,A,\delta \rangle $ \cite{modi2005adopt} where,
\begin{itemize}
    \item A is a set of agents $\{a_1, a_2, ..., a_n\}$.
    \item X is a set of discrete variables $\{x_1, x_2, ..., x_m\}$, which are being controlled by the set of agents A.
    \item D is a set of discrete and finite variable domains $\{D_1, D_2, ..., D_m\}$, where each $D_i$ is a set containing values which may be assigned to its associated variable $x_i$.
    \item F is a set of constraints $\{f_1,f_2,...,f_l\}$, where $f_i \in F$ is a function of a subset of variables $x^i \subseteq X$ defining the relationship among the variables in $x^i$. Thus, the function $f_i : \times_{x_j \in x^i} D_j \to \!R $ denotes the cost for each possible assignment of the variables in $x^i$. 
    \item $\delta: X \rightarrow
	A$ is a variable-to-agent mapping function \cite{nodeto} which assigns the control of each variable $x_i \in X$ to an agent of $A$. Each variable is controlled by a single agent. However, each agent can hold several variables. 
\end{itemize}
Within the framework, the objective of a DCOP algorithm is to produce $X^*$; a complete assignment that minimizes\footnote{For a maximization problem $\argmin$ is replaced with $\argmax$ in Equation~\ref{eqobj}.} the aggregated cost of the constraints as shown in Equation~\ref{eqobj}.

\begin{equation}
    X^* = \argmin_X \sum_{i=1}^{l} f_i(x^i)
    \label{eqobj}
\end{equation}
For ease of understanding, we assume that each agent controls one variable. Thus, the terms `variable' and `agent' are used interchangeably throughout this paper. 

\subsection{Distributed Simulated Annealing}
\noindent Distributed Simulated Annealing (DSAN) \cite{ArshadDistributedSA} is the only existing Simulated Annealing (SA) based DCOP solver. DSAN is a local search algorithm that executes the following steps iteratively:
\begin{itemize}
    \item Each agent $a_i$ selects a random value $d_j$ from domain $D_i$.
    \item Agent $a_i$ then assigns the selected value to $x_i$ with the probability $min(1,exp(\frac{\Delta}{t_i}))$ where $\Delta$ is the local improvement if $d_j$ is assigned and $t_i$ is the temperature at iteration $i$. Note that the authors of DSAN suggest that the value of $t_i = \frac{Max\_Iteration}{i^2}$ or $t_i = \frac{1}{i^2}$. However, setting the value of the temperature parameter with such a fixed schedule does not take into account their impact on the performance of the algorithm. 
    \item Agents notify neighbouring agents if the value of a variable changes.
\end{itemize}
\subsection{Anytime Local Search}
\noindent Anytime Local Search (ALS) is a general framework that can be used to give distributed iterative local search DCOP algorithms such as DSAN (described above), DSA, GDBA an anytime property. Specifically, ALS uses a BFS-tree to calculate the global cost (i.e. evaluate Equation \ref{eqobj}) of the system's state during each iteration and keeps track of the best state visited by the algorithm. Using this framework, agents can use the best assignment decision that they explored during the iterative search process instead of the one that occurs at the termination of the algorithm (see  \cite{zivan2014explorative} for more details).   
\subsection{Cross-Entropy Sampling}
\begin{algorithm}[t]
\DontPrintSemicolon
Initialize parameter vector $\theta$, $\#S$, $G$,  $\alpha$\;
\While {condition not met}{
    $X \leftarrow$ take $\#S$ samples from distribution $\mathcal{G}$($\theta$)\;
    $S \leftarrow$ Evaluate points in $X$ on the objective\;
    $X \leftarrow$ sort($X$, $S$)\;       
    $\theta_{new} \leftarrow$ calculate updated $\theta$ using X(1:G)\;
    $\theta \leftarrow $ $(1-\alpha)*\theta+\alpha*\theta_{new}$
}
\caption{Cross-Entropy Sampling}
\label{algo:CEM}
\end{algorithm}
\noindent Cross-Entropy (CE) is a Monte Carlo method for importance sampling. CE has successfully been applied to importance sampling, rare-event simulation and optimization (discrete, continuous, and noisy problems) \cite{Kroese2011HandbookOM}. Algorithm \ref{algo:CEM} sketches an example that iteratively searches for the optimal value of the $X$ within a search space. The algorithm starts with a probability distribution $\mathcal{G}(\theta)$ over the search space with parameter vector $\theta$ initialized to a certain value (that may be random). At each iteration, it takes \#$S$ (which is a parameter of the algorithm) samples from the probability distribution $\mathcal{G}$($\theta$) (line $3$). After that, each sample point is evaluated on a problem-dependent objective function. The top $G$ among the \#$S$ sample points are used to calculate the new value of $\theta$ which is referred to as $\theta_{new}$ (lines $4-6$). Finally, $\theta_{new}$ is used to update $\theta$ (line $7$). At the end of the learning process, most of the probability density of $\mathcal{G}(\theta)$ will be allocated near the optimal value of\:$X$.    
\section{The AED Algorithm}
\noindent Anytime Evolutionary DCOP (AED) is a synchronous iterative population-based algorithm that adapts an evolutionary process to solve DCOPs. More precisely, AED starts with a population of randomly generated candidate solutions (individuals) and each agent is responsible for maintaining a subset of the population (local population). At each iteration, new individuals are added to the population by modifying current individuals locally (reproduction). Following that, a stochastic selection process takes place at each local population where a subset of the most promising individuals are kept and the rest are discarded. This is interpreted as an evolutionary process. Besides producing better solutions than state-of-the-art algorithms (see Section 6), one advantage AED has over ACO\_DCOP is that AED only requires communication between neighbouring agents whereas ACO\_DCOP requires broadcasting (each agent communicating with all the other). This is a significant disadvantage for ACO\_DCOP since many applications of DCOPs are limited to point-to-point communication (e.g. applications related to sensor networks). In this section, we first present the AED algorithm, then prove it is anytime and finally provide a complexity analysis.          
\subsection{The Proposed Method}
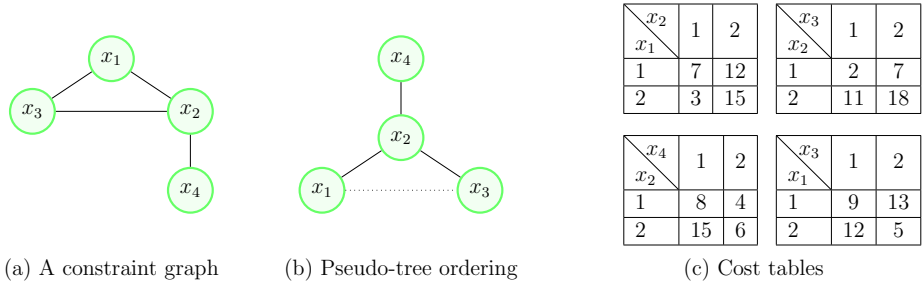
\begin{figure}[t]
\vspace{-10mm}
\centering
\scalebox{.7}{
  \begin{tikzpicture}
        [
        roundnode/.style={circle, draw=green!60, fill=green!5, very thick, minimum size=7mm},
        ]
        \node[roundnode]    at(-6, 0)  (x1)         {$x_1$};
        \node[roundnode]    at(-4.5,-1)  (x2)      {$x_2$};
        \node[roundnode]    at(-7.5,-1)  (x3)      {$x_3$};
        \node[roundnode]    at(-4.5,-2.5)  (x4)      {$x_4$};
         
        \draw (x1) -- (x2);
        \draw (x1) -- (x3);
        \draw (x3) -- (x2);
        \draw (x4) -- (x2);
        \node  at (-6,-4)
        {
            (a) A constraint graph
        };
        \node at (5,0)  {
            \begin{tabular}{|l|c|c|}\hline
            \diagbox[width=2.5em]{$x_1$}{$x_2$}&
              1 & 2 \\ \hline
            1 & 7 & 12 \\ \hline
            2 & 3 & 15 \\ \hline
            \end{tabular}
        };
        \node at (8,0)  {
            \begin{tabular}{|l|c|c|}\hline
            \diagbox[width=2.5em]{$x_2$}{$x_3$}&
              1 & 2 \\ \hline
            1 & 2 & 7 \\ \hline
            2 & 11 & 18 \\ \hline
            \end{tabular}
        };
        \node at (5,-2.5)  {
            \begin{tabular}{|l|c|c|}\hline
            \diagbox[width=2.5em]{$x_2$}{$x_4$}&
              1 & 2 \\ \hline
            1 & 8 & 4 \\ \hline
            2 & 15 & 6 \\ \hline
            \end{tabular}
        };
        \node at (8,-2.5)  {
            \begin{tabular}{|l|c|c|}\hline
            \diagbox[width=2.5em]{$x_1$}{$x_3$}&
              1 & 2 \\ \hline
            1 & 9 & 13 \\ \hline
            2 & 12 & 5 \\ \hline
            \end{tabular}
        };
        \node  at (6.2,-4)
        {
            (c) Cost tables 
        };
        [
        roundnode/.style={circle, draw=green!60, fill=green!5, very thick, minimum size=7mm},
        ]
        \node[roundnode]    at(-0.5,0)  (x4)                              {$x_4$};
        \node[roundnode]    at(-0.5,-1.5)  (x2)                  {$x_2$};
        \node[roundnode]    at(-2,-2.5)  (x1)                              {$x_1$};
        \node[roundnode]    at(1,-2.5)  (x3)                              {$x_3$};
        \draw (x4) -- (x2);
        \draw (x2) -- (x3);
        \draw (x2) -- (x1);
        \draw[dotted] (x1) -- (x3);
         \node  at (-0.5,-4)
        {
            (b) Pseudo-tree ordering 
        };
    \end{tikzpicture}
}
\vspace{-2mm}
\caption{A sample DCOP instance.}
\label{dcopex}
\end{figure}

\noindent AED consists of two phases: Initialization (Algorithm~\ref{algo:AED}: lines 1-2) and Optimization (Algorithm~\ref{algo:AED}: lines 3-13). During the former, agents initially order themselves into a pseudo-tree (Algorithm~\ref{algo:AED}: lines 1), then initialize the necessary variables and parameters (Procedure 1: lines 2). Finally, they make a random assignment to the variables they control (Procedure 1: lines 3) and cooperatively construct the initial population (Procedure 1: lines 4-17). During the latter phase (Algorithm~\ref{algo:AED}: lines 3-11), agents iteratively improve this initial population. To do this, a reproduction step takes place (Algorithm~\ref{algo:AED}: lines 4-5) that constructs new individuals using individuals from the current population. These newly created individuals then get added back to the local population and the entire population goes through a stochastic selection process where a subset of the most promising individuals are kept and the rest are discarded (Algorithm ~\ref{algo:AED}: lines 7-8). Finally, agents exchange individuals from their population with their neighbours (Algorithm~\ref{algo:AED}: lines 10-13). This plays an important role in global optimization since reproduction only locally optimizes individuals. During this entire process, an agent also keeps track of the best individual using our proposed anytime update mechanism (Algorithm~\ref{algo:AED}: line 9 and Procedure 3). We will now discuss both of the phases along with the procedures in more detail. Throughout this process, we will use Figure~\ref{dcopex} to provide examples. Figure~\ref{dcopex}a illustrates a sample DCOP using a constraint graph where each node represents an agent $a_i \in A$ labelled by a variable $x_i \in X$ that it controls and each edge represents a function $f_i \in F$ connecting all $x_j \in x^i$. Figure~\ref{dcopex}c shows the corresponding cost tables. \par

\begin{algorithm}[!ht]
\DontPrintSemicolon
\small
   Construct pseudo-tree\;
   Every agent $a_i$ calls INIT( )\;
   \While{Stop condition not met each agent $a_i$}
   {
        $P_{new} \leftarrow P_{a_i}$ \; 
   		$\forall I \in P_{new}$, Modify individual I by Equations~\ref{eqw1},~\ref{eqw2},~\ref{eqw3},~\ref{eqdel}\;
   		$P_{a_i} \leftarrow P_{a_i} \cup P_{new}$ \;
   		$B \leftarrow \argmin_{I \in P_{a_i}} I.fitness$\;
   		$P_{a_i} \leftarrow Select(P_{a_i},|N_i|*ER)$ \;
   		Anytime Update$(B)$\;
   		$P_{send} \leftarrow$ Partition $P_{a_i}$ into equal size subsets randomly $\{P_{send}^{n_1},...,P_{send}^{n_{|N_i|}}\}$\;
   	    $\forall n_j \in N_i$ Send $P_{send}^{n_j}$ to $a_{n_j}$\;
   	    $P_{a_i} \leftarrow \emptyset$\;
        $P_{received}^{n_{j}}$ received from $\forall n_j \in N_i$, $P_{a_i} \leftarrow P_{a_i} \cup P_{received}^{n_{j}}$\;
      
   }
\caption{Anytime Evolutionary DCOP}
\label{algo:AED}
\end{algorithm}

\textbf{The Initialization Phase} of AED  consists of two parts: pseudo-tree construction and running INIT($\cdot$) (Procedure 1) that initializes the algorithm parameters, variables and population (Algorithm~\ref{algo:AED}: Line 1-2). This phase starts by ordering the agents into a Pseudo-Tree. This ordering serves two purposes. It helps in the construction of the initial population and facilitates Anytime Update (Procedure 3) during the optimization phase. Even though either of the BFS or DFS pseudo-tree can be used, AED uses a BFS Pseudo-tree\footnote{A BFS Pseudo-tree can be constructed in a distributed manner using \cite{hen2017improved}.}. This is because it generally produces a pseudo-tree with smaller height\footnote{Length of the longest path in the pseudo-tree.} \cite{zivan2014explorative}, which improves the performance of Anytime Update (see Theoretical Analysis for details). Figure~\ref{dcopex}b shows an example of a BFS pseudo-tree constructed from the constraint graph shown in Figure~\ref{dcopex}a having $x_4$ as the root. Here, the height (i.e. H = 2) of this pseudo-tree is calculated during the time of construction and is maintained by all agents. From this point, $N_i$ refers to the set of neighbours; $C_i \subseteq N_i$ refers to the set of child nodes and ${PR}_i$ refers to the parent of an agent $a_i$ in the pseudo-tree. For instance, we can see in Figure~\ref{dcopex}b that $N_2 = \{a_1,a_3,a_4\}$, $C_2 = \{a_1,a_3\}$ and ${PR}_2 = a_4$ for agent $a_2$. After the pseudo-tree construction, all the agents synchronously call the procedure INIT($\cdot$) (Algorithm~\ref{algo:AED}: Line 2).\par

INIT($\cdot$) starts by initializing all the parameters and variables to their default values\footnote{AED takes a default value for each of the parameters as input. Default values of the variables are discussed later in this section.}.
Then each agent $a_i$ sets its variable $x_i$ to a random value from its domain $D_i$. Lines 3 to 17 of Procedure 1 describe the initial population construction process. In AED, we define population \textit{P} as a set of individuals that are collectively maintained by all the agents and the local population $P_{a_i} \subseteq P$ as the subset of the population maintained by agent $a_i$.
An individual in AED is represented by a complete assignment of variables in $X$ and the fitness\footnote{a value that defines how good an individual is.} is calculated using the function shown in Equation~\ref{eqfit}. 
This function calculates the aggregated cost of constraints yielded by the assignment. Hence, optimizing this fitness function results in an optimal solution for the corresponding DCOP.\par 

\begin{equation}
    fitness = \sum_{f_i\in F}^{} f_i(x^i) 
    \label{eqfit}
\end{equation}

\begin{procedure}[!ht]
  \caption{INIT()( )}
  \DontPrintSemicolon
  \small
    Initialize parameters IN, ER, $\alpha, \beta$ and variables LB, GB, FM, UM\;
    $x_i \leftarrow$ random value from $D_i$\;
    $P_{a_i} \leftarrow $Set of empty individuals of size IN\;
    $\forall I \in P_{a_i}$, $I.x_i \leftarrow$ Choose\_Random($D_i$)\;
    Send $P_{a_i}$ to agents in $N_i$  \;
    $P_{n_j}$ received from $\forall n_j \in N_i$, $P_{a_i} \leftarrow Merge(P_{a_i},P_{n_j})$\;
    $\forall I \in P_{a_i}$, $I.fitness \leftarrow \sum_{n_j\in N_i}^{} Cost_{i,j}(I.x_i,I.x_j) $\;
    Wait until received $P_{c_j}$ from all $c_j \in C_i$\;
    $P_{c_j}$ received from $\forall c_j \in C_i$, $P_{a_i} \leftarrow Merge(P_{a_i},P_{c_j})$\;
    \eIf{$a_i \not= root$}
    {
        Send $P_{a_i}$ to ${PR}_i$\;
    }
    {
        $\forall I \in P_{a_i}$, $I.fitness \leftarrow I.fitness/2 $\;
        Send $P_{a_i}$ to all agent in $C_i$\;
    } 
    \If{Received $P_{{PR}_i}$ from ${PR}_i$}{
    $P_{a_i} \leftarrow P_{{PR}_i}$\;
    Send $P_{a_i}$ to all agent in $C_i$\;
    }
    \label{initp}
\end{procedure}
Note that a single agent cannot calculate the fitness function since it does not have the knowledge about all the constraints. In AED, it is calculated in parts with the cooperation of all the agents during the construction process. Moreover, the fitness value is added to the representation of an individual because it enables an agent to recalculate the fitness when a new individual is constructed during the reproduction processes only using local information. We take I = $\{x_1=1, x_2=2, x_3=1, x_4=2,fitness = 38\}$ as an example of a complete individual from the DCOP shown in Figure~\ref{dcopex}. We use the dot(.) notation to refer to a specific element of an individual. For example $I.x_1$ refers to $x_1$ in the individual I. Additionally, we define a Merger operation of two individuals under construction, $I_1$ and $I_2$ as Merge($I_1, I_2$). This operation constructs a new individual $I_3$ by aggregating the assignments and setting $I_3.fitness = I_1.fitness+I_2.fitness$. We define an extended Merge operation for two ordered sets of individuals $S_1$ and $S_2$ as $Merge(S_1,S_2) = \{I_i: Merge(S_1.I_i,S_2.I_i)\}$ where $I_i$ is the i-th individual in a set. \par

At the beginning of the construction process, each agent $a_i$ sets $P_{a_i}$ to a set of empty individuals\footnote{Individuals with no assignment and fitness set to 0.}. The size of the initial $P_{a_i}$ is defined by parameter IN. Then for each individual $I \in P_{a_i}$, agent $a_i$ makes a random assignment to $I.x_i$. After that, each agent $a_i$ executes a merger operation on $P_{a_i}$ with each local population maintained by agents in $N_i$ (Procedure 1: Lines 2-6). At this point, an individual $I \in P_{a_i}$ consists of an assignment of variables controlled by $a_i$, and agents in $N_i$ with fitness set to zero. For example, I = $\{x_1=1, x_2=2, x_3=1, fitness = 0\}$ represents an individual of $P_{a_3}$. The fitness of each individual is then set to the local cost according to their current assignment (Procedure 1: Line 7). Hence, the individual I from the previous example becomes $\{x_1=1, x_2=2, x_3=1, fitness = 20\}$. In the next step, each agent $a_i$ executes a merger operation on $P_{a_i}$ with each local population that is maintained by the agents in $C_i$. Then each agent $a_i$ sends $P_{a_i}$ to ${PR}_i$ apart from the root (Procedure 1: Line 8-11). At the end of this step, the local population maintained by the root consists of complete individuals. However, their fitness is twice its actual value since each constraint is calculated twice. Therefore, the root agent at this stage corrects all the fitness values (Procedure 1: Lines 13). Finally, the local population of the root agent is distributed through the network so that agents can initialize their local population (Procedure 1: Line 14-17). This concludes the initialization phase and all the agents can now synchronously start the optimization phase to improve this initial population iteratively.\par  
 
\textbf{The Optimization Phase} of AED consists of four steps, namely Reproduction, Reinsertion, Anytime Update and Migration that are applied iteratively. An agent $a_i$ begins each iteration by constructing new individuals from current individuals of the local population (Algorithm~\ref{algo:AED}: lines 4-5). In order to do this, each individual $I$ of the local population $P_{a_i}$ of agent $a_i$ is mutated using Equations ~\ref{eqw1},~\ref{eqw2} and ~\ref{eqw3} at the position $I.x_i$ (Algorithm~\ref{algo:AED}: line 5). This mutation process begins by calculating the cost caused by each possible mutation using Equation ~\ref{eqw1}. After that, the advantage of each possible mutation is calculated by subtracting the worst mutation cost using Equation~\ref{eqw2}. This equation is normalized to keep the advantage between $(0,1]$. Finally, the mutation probability; on which the mutation is based is calculated using Equation~\ref{eqw3}. In Equation~\ref{eqw3}, $\beta$ is used to balance between exploration and exploitation. With the increase of $\beta$, the selection curve gets steeper. The $\epsilon$ is added in Equation~\ref{eqw2} to allocate a small probability to even the worst mutation. Additionally, agents $a_i$ calculates the new fitness after mutation of individual $I$ by adding $\delta$ to I.fitness. Here, $\delta$ is calculated using Equation ~\ref{eqdel} where $I.x_i^{new}$ and $I.x_i^{old}$ are the old and new values of $I.x_i$, respectively.\par

\begin{equation}
    MutationCost_{d_j} = \sum_{n_k \in N_i}Cost_{i,k}(I.x_i = d_j,I.x_{n_k})
    \label{eqw1}
\end{equation}
\begin{equation}
    MutationAdvantage_{d_j} = \frac{|MutationCost_{worst} - MutationCost_{d_j}|+\epsilon}{|MutationCost_{worst} - MutationCost_{best}|+\epsilon}
    \label{eqw2}
\end{equation}
\begin{equation}
    P(d_j) = \frac{MutationAdvantage_{d_j}^{\beta}}{\sum_{d_k \in D_i}MutationAdvantage_{d_k}^{\beta}}
    \label{eqw3}
\end{equation}
 \begin{equation}
    \delta = \sum_{n_k \in N_i}Cost_{i,k}(I.x_{i}^{new},I.x_{n_k})-Cost_{i,k}(I.x_i^{old},I.x_{n_k})
    \label{eqdel}
\end{equation}

To illustrate an example of this Reproduction mechanism, consider agent $a_3$ of Figure~\ref{dcopex} that wants to creates a new individual by mutating $I = \{x_1=1, x_2=2, x_3=2, x_4=2,fitness = 49\}$. Here, the domain of agent $a_3$ is $\{1,2\}$. At $I.x_{3}$, there are two possible mutations $I.x_{3} = 1$ and $I.x_{3} = 2$. Using Equation~\ref{eqw1}, $MutationCost_{x_3=1} = 9$ and $MutationCost_{x_3=2} = 20$. Consider $\epsilon = 1$, then using Equation~\ref{eqw2}, $MutationAdvantage_{x_3=1} = \frac{11+1}{11+1} = 1$ and $MutationAdvantage_{x_3=2} = \frac{0+1}{11+1} = 0.083$. Now, if $\beta = 1$, mutation probability $P({x_3=1}) = 0.923$ and $P({x_3=2}) = 0.0777$. If the $\beta$ were 3 it would have been, $P(x_3=1) = 0.999$ and $P(x_3=2) = 0.001$. Hence, a high value of $\beta$ means more exploitation and less exploration. Suppose after mutation $I.x_i =1$, then the fitness is updated by adding $\delta = -11$ to I.fitness.\par 

After Reproduction, agent $a_i$ adds all the new individuals to the local population $P_{a_i}$ which doubles its size. To keep the population size bounded a Reinsertion step takes place (Algorithm~\ref{algo:AED}: Lines 6,8) where agents select half of the population for the next iteration and discard the rest. There are several possible ways to select individuals for this purpose that has been studied in the evolutionary optimization literature. In AED, we explore a fitness proportionate selection scheme which is one of the most widely used selection schemes in the literature \cite{Blickle1996ACO, hancock1994empirical}. According to this scheme, first, all the individuals are scored from $(0, 1]$ based on their relative fitness in the local population $P_{a_i}$. The rank $Rank_j$ of an individual $I_j \in P_{a_i}$ is calculated using Equation~\ref{eqs1}. Here, $I_{best}$ and $I_{worst}$ are the individuals with the lowest and highest fitness in $P_{a_i}$ respectively\footnote{For minimization problems, a lower value of fitness is better.}. Then individuals are sampled with replacement\footnote{Any individual can be selected more than once.} from population $P_{a_i}$ based on the probability calculated using Equation ~\ref{eqs2}. In Equation~\ref{eqs2}, $\alpha$ contributes to the exploration and exploitation dynamics in a similar manner as $\beta$ did in Equation~\ref{eqw3}. For example, assume $P_{a_i}$ consists of 3 individuals $I_1, I_2, I_3$ with fitness 16, 30, 40 respectively and $\epsilon = 1$. Then Equations ~\ref{eqs1} and ~\ref{eqs2} will yield, $P(I_1)=0.676, P(I_2)=0.297, P(I_3)=0.027$ if $\alpha = 1$ and $P(I_1)=0.92153, P(I_2)=0.07842, P(I_3)=0.00005$ if $\alpha = 3$.\par 
\begin{equation}
    Rank_j =  \frac{|I_{worst}.fitness - I_j.fitness|+\epsilon}{|I_{worst}.fitness - I_{best}.fitness|+\epsilon}
    \label{eqs1}
\end{equation}
\begin{equation}
    P(I_j) = \frac{Rank_{j}^{\alpha}}{\sum_{I_k \in P_{a_i}} Rank_k^{\alpha}}
    \label{eqs2}
\end{equation}\par
 
\begin{procedure}[!ht]
  \caption{Anytime Update($B$)}
    \DontPrintSemicolon
    \small
   \If{$B.fitness < LB.fitness$}{
        $LB \leftarrow B$\;
   }
   \If{$LB.fitness < GB^{Itr}.fitness$}{
        \eIf{$a_i = root$}{
            $GB^{itr} \leftarrow LB$\;
            $UM \leftarrow \{V:Itr,I:LB\}$
        }{
            $FM \leftarrow \{I:LB\}$
        }
   }
   Send Update message UM to agents $\in C_i$ if $UM \not= \emptyset$\; 
   Found message FM to ${PR}_i$ if $FM \not= \emptyset$\;
   set $FM$ and $UM$ to $\emptyset$\;
   \If{Received Update message M}{
        $GB^{M.V} \leftarrow M.I$\;
        $LB \leftarrow $Best between LB and M.I\;
        $UM \leftarrow M$\;
   }
   \If{Received Found message M and and $M.I.fitness < LB.fitness$}{
        $LB \leftarrow M.I$\;
   }
   \If{$Itr >= H$}{
            $x_i = GB^{Itr-H+1}.x_i$\;
   }
   \label{anytimep}
\end{procedure}
Following the Reinsertion, the Anytime Update step takes place. This helps the agents to keep track of the best individuals in the entire population and to update their assignment according to that. To facilitate the anytime update mechanism, each agent maintains four variables $LB, GB, FM, UM$. $LB$ (Local Best) and $GB$ (Global Best) are initialized to empty individuals with fitness set to infinity. $FM$ and $UM$ are initialized to $\emptyset$. Additionally, $GB$ is stored with a version tag and each agent maintains previous versions of $GB$ having version tags in the range $[Itr-H+1, Itr]$ (see the Theoretical section for details). Here, $Itr$ refers to the current iteration number. We use $GB^j$ to refer to the latest version of $GB$ with the version tag not exceeding $j$. Our proposed anytime update mechanism works as follows. Each agent keeps track of two different bests, $LB$ and $GB$. Whenever the fitness of LB becomes less than $GB$\footnote{when minimizing cost.}, it has the potential to be the global best solution. So it gets reported to the root through the propagation of a Found message up the pseudo-tree. Since the root gets reports from all the agents, it can identify the true global best solution, and notify all the agents by propagating an Update message down to the pseudo tree. The root also adds the version tag ($V$) in the Update message to help coordinate variable assignment. Next, Anytime Update starts by keeping $LB$ updated with the best individual $B$ in $P_{a_i}$. In line 3 of Procedure 3, agents try to identify whether $LB$ is the potential global best. When identified, and if the identifying agent is the root, it is the true global best and an Update message $UM$ is constructed. If the agent is not the root, it is a potential global best and a Found message $FM$ is constructed (Procedure 3: Lines 4-8). Each agent forwards the message $UM$ to agents in $C_i$ and the message $FM$ to the ${PR}_i$. Upon receiving these messages, an agent takes the following actions: 
\begin{itemize}
    \item If an Update message is received then an agent updates both its $GB$ and $LB$. Additionally, the agent saves the Update message in $UM$ and sends it to all the agents in $C_i$ during the next iteration (Procedure 2: Lines 12-15). 
    \item If a Found message is received and it is better than $LB$, only $LB$ is updated. If this remains a potential global best it will be sent to ${PR}_i$ during the next iteration (Procedure 2: Lines 16-17).
\end{itemize}\par

An agent $a_i$ then updates the assignment of $x_i$ using $GB^{Itr-H+1}$ (Procedure 2: Lines 18-19). Agents make decisions based on $GB^{Itr-H+1}$ instead of the potentially newer $GB^{Itr}$ so that decisions are made based on the same version of $GB$. $GB^{Itr-H+1}$ will be same for all agents since it takes at most $H$ iterations for an Update message to propagate to all the agents. For example, assume agent $a_1$ from Figure~\ref{dcopex}b finds a potential best individual $I$ at $Itr = 3$. Unless it gets replaced by a better individual, it will reach the root $a_4$ via agent $a_2$ through a Found message at $Itr =4$. Then $a_4$ constructs an Update message $\{Version: 5, Individual:I\}$ at $Itr=5$. This message will reach all the agents by $Itr=6$ and the agents save it as $GB^5=I$. Finally, at $Itr=6$ agents assign their variables using $GB^{6-2+1}=GB^{5}$ which is the best individual found at $Itr=3$.\par   

Finally, Migration, an essential step of AED, takes place. We outline this in lines 9-11 of Algorithm ~\ref{algo:AED}. For this step, Migration is a simple process of exchanging individuals among the neighbours. In AED, the Reproduction mechanism performs mutation only on the local variables. This means only a subset of the variables of an individual are changed. However, because of Migration, different agents can change a different subset of variables as individuals get to traverse the network through this mechanism. Hence, this step plays an essential role in the optimization process of AED. During this step, an agent $a_i$ divides its local population $P_{a_i}$, into $|N_i|$ subsets of size ER randomly and exchanges those individuals with the neighbours. Upon collecting individuals from all the neighbours, an agent $a_i$ adds them to its local population $P_{a_i}$. This concludes an iteration of the optimization phase and every step repeats during the subsequent iterations.\par

\subsection{Theoretical Analysis}
\noindent In this section, we first prove that AED is anytime, that is the quality of solutions found by AED increase monotonically. Then we analyze the complexity of AED in terms of communication, computation and memory requirements.
\newtheorem{thm}{Theorem}
\newtheorem{lem}[thm]{Lemma}
\newtheorem{prp}[thm]{Proposition}
\newproof{pf}{Proof}
\newtheorem{pfl}[pf]{Proof of Lemma}
\newtheorem{pfp}[pf]{Proof of Proposition}
\begin{lem}
At iteration $\mathbf{i+H}$, the root agent is aware of the best individual in \textit{P} at least up to iteration $\mathbf{i}$.
\label{lem1}
\end{lem}
 
\begin{pfl}
Suppose, the best individual up to iteration $\mathbf{i}$ is found at iteration $\mathbf{i^\prime}\le \mathbf{i}$ by agent $\mathbf{a_x}$ at level $\mathbf{l^\prime}$. Afterwards, one of the following 2 cases will occur at each iteration.
\begin{itemize}
    \item Case 1. This individual will be reported to the parent of the current agent through a Found message.
    \item Case 2. This individual gets replaced by a better individual on its way to the root at iteration $\mathbf{i^{*}}>\mathbf{i^\prime}$ by agent $\mathbf{a_{y}}$ at level\:$\mathbf{l^{*}.}$
\end{itemize}
When only Case 1 occurs, the individual will reach the root at iteration $\mathbf{i^{\prime}}+\mathbf{l^{\prime}} \le \mathbf{i}+\mathbf{H}$ (since ${l^{\prime}}$ can be at most H). If Case 2 occurs, the replaced individual will reach the root agent by $\mathbf{i^*}+\mathbf{l^{*}} = \{\mathbf{i^*} - (\mathbf{l^{\prime}}-\mathbf{l^{*}}) \}+ \{(\mathbf{l^{\prime}}-\mathbf{l^{*}}) + \mathbf{l^{*}}\} = \mathbf{i^{\prime}}+\mathbf{l^{\prime}} \le \mathbf{i}+\mathbf{H}$. The same can be shown when the new individual also gets replaced. In either case, at iteration \textbf{i+H}, the root will become aware of the best individual in \textit{P} up to iteration \textbf{i} or will become aware of a better individual in \textit{P} found at iteration $\mathbf{i^{*}}>\mathbf{i}$; meaning the root will be aware of the best individual in \textit{P} at least up to iteration \textbf{i}. 
\end{pfl}
\begin{lem}
The variable assignment decision made by all the agents at iteration $\mathbf{i+2H-1}$ yields a global cost equal to the fitness of the best individual in \textit{P} at least up to iteration $\mathbf{i}$.
\label{lem2}
\end{lem}
\begin{pfl}
At iteration $\mathbf{i+2H-1}$, all the agents make decisions about variable assignment using $\mathbf{GB^{i+H}}$. However, $\mathbf{GB^{i+H}}$ is the best individual known to the root up to iteration $\mathbf{i+H}$. We know from Lemma~\ref{lem1} that, at iteration $\mathbf{i+H}$, the root is aware of the best individual in \textit{P} at least up to iteration $\mathbf{i}$. Hence, the fitness of $\mathbf{GB^{i+H}}$ is at least equal to the best individual in \textit{P} up to iteration $\mathbf{i}$. Hence, at iteration $\mathbf{i+2H-1}$, it yields a global cost equal to the fitness of the best individual in \textit{P} at least up to iteration $\mathbf{i}$.
\end{pfl}
\begin{prp}
AED is anytime.
\end{prp}
\begin{pfp}
From Lemma~\ref{lem2}, the decisions regarding the variable assignments at iterations $\mathbf{i+2H-1}$ and $\mathbf{i+2H-1+\delta}$ yields a global cost equal to the fitness of the best individual in \textit{P} at least up to iterations $\mathbf{i}$ and  $\mathbf{i+\delta}$ ($\mathbf{\delta \ge 0}$), respectively. Now, the fitness of the best individual in \textit{P} up to iteration $\mathbf{i+\delta}$ is at most the fitness at iteration $\mathbf{i}$. So the global cost at iteration $\mathbf{i+\delta}$ is less than or equal to the same cost at iteration $\mathbf{i}$. As a consequence, the quality of the solution monotonically improves as the number of iterations increases. Hence, AED is anytime.   
\end{pfp}
We now consider algorithm complexity. Assume, \textbf{n} is the number of agents, $\mathbf{|N|}$ is the number of neighbours and $\mathbf{|D|}$ is the domain size of an agent. In every iteration, at most $|N|$ messages are passed for each of the Anytime Update and Migration steps. Now, $|N|$ can be at most n (complete graph). Hence, the total number of messages transmitted per agent during an iteration is $O(2|N|) = O(n)$. Since the main component of a message in AED is the set of individuals, the size of a single message can be calculated as the size of an individual multiplied by the number of individuals. During the Reproduction, Migration and Anytime Update steps, at most ER individuals, each of which has size $O(n)$, is sent in a single message. As a result, the size of a single message is $O(ER*n)$. This makes the total message size per agent during an iteration $O(ER*n*n) = O(ER*n^2)$.\par
Before Reproduction, $|P_{a_i}|$ is $ER*|N|$ and Reproduction will add $ER*|N|$ individuals. So the memory requirement per agent is $O(2*ER*|N|*n) = O(n^2)$. Finally, Reproduction using Equations ~\ref{eqw1},~\ref{eqw2},~\ref{eqw3} and ~\ref{eqdel} requires $|D_i|*|N|$ operations and in total $ER*|N|$ individuals are reproduced during an iteration per agent. Hence, the total computation complexity per agent during an iteration is $O(ER*|N|*|D|*|N|) = O(ER*|D|*n^2)$.\par 

\section{The DPSA Algorithm}
\noindent The DPSA is a hybrid population-based algorithm based upon the Simulated Annealing (SA) meta-heuristic. One of the most important factors that influence the quality of the solution produced by SA is its temperature parameter, widely denoted as $T$. More precisely, SA starts with a high value of T and during the search process continuously cools it down to near zero. When T is high, SA only explores the search space without exploiting. This makes its behaviour similar to a random search procedure. On the other hand, when T is near zero, SA tends to only exploit and thus the exploration capability decreases. In such a scenario, SA emulates the behaviour of a greedy algorithm. In fact, SA most effectively balances between exploration and exploitation in some optimal temperature region that lies in between these two extremes. Several existing works also discuss a constant optimal temperature where SA performs the best \cite{CONNOLLY199093, alrefaei1999simulated}. Unfortunately, the optimal temperature region varies from one type of problem to another and from one instance to another of the same type problem (e.g. with different constraint functions and constraint densities).\par
Based on the above observation, we develop DPSA that keeps a set of systems\footnote{A system is a candidate solution of the given problem (i.e. a complete assignment). Each system is maintained similarly as a local search DCOP algorithm (e.g. DSAN, DSA) maintains a single solution. A system in DPSA is similar to an Individual in AED or an Ant in ACO\_DCOP.} each of which gets modified by applying Distributed Simulated Annealing (DSAN) at a different constant temperature. Based on the performance yielded by these systems at a different temperature, agents try to learn an optimal temperature region using a Monte Carlo importance sampling method called Cross-Entropy sampling. Using the knowledge learned during this process, agents also cooperatively solve the given problem. This results in a significant improvement in the solution quality produced by the algorithm (see Section 6). Further, DPSA is more scalable than the existing population-based algorithms (as we will see in Section 6). This is because, in existing population-based algorithms, agents communicate entire populations between themselves. This limits scalability and produces a large network load. DPSA avoids this by maintaining each of the systems in a distributed manner similar to local search algorithms and only communicate the assignment of a single variable rather than a complete assignment. It is worth mentioning that the success of the DPSA depends on the prior knowledge about the search space of the temperature parameter (explained in detail in subsequent subsections). However, such knowledge may not always be available. Anticipating this, we also provide a novel pruning algorithm, called Greedy Baseline, that helps alleviate this requirement. The rest of the section is organised as follows. First, we describe the DPSA algorithm in detail. We then present the Greedy Baseline method. Finally, we analyze the complexity of DPSA.\par

\subsection{Proposed Method}

\noindent As discussed earlier, a big motivation behind DPSA (Algorithm $3$) is to learn the optimal temperature region for Simulated Annealing (SA). Thus, it is important that we formally define both optimal temperature (Definition 1) and optimal temperature region (Definition 2).\par
\textbf{Definition\:1.} \textit{An Optimal Temperature given simulation length $L$} is a constant temperature at which the expected solution cost yielded by running SA for L iterations is the lowest of all the temperatures $> 0$.\par   
\textbf{Definition 2:} \textit{An Optimal Temperature Region (OTR) of length $\epsilon$} is a continuous interval $[T_{min},T_{max}]$ where $T_{max} - T_{min} \le \epsilon$ and contains the optimal temperature. If we set $T_{min}$ to near zero and $T_{max}$ to a very large number, it will always be an OTR by the above definition; although not a useful one. The proposed DPSA algorithm tries to find an OTR with sufficiently small\footnote{defined on the input of Algorithm 3.} $\epsilon$.\par

\begin{algorithm}[t]
\DontPrintSemicolon
\small
\SetKwFunction{Round}{Simulate}
\SetKwProg{In}{Function}{:}{} 
       Construct BFS Tree\;
       Initialize parameters: Parameter Vector $\theta$, $Itr_{max}, R_{max}, S_{max}, S_{len}, G, K$\;
        \For{$R = 1 ... R_{max}$ AND Conditions are met}{
            $E \leftarrow \{e_1 = 0,e_2 = 0,...,e_K = 0\}$\;
            \eIf{the agent is the root}
           	    {
           	        $T \leftarrow \{t_1,t_2,...,t_K\}$sampled from $\mathcal{G}(\theta)$\;
           	    }
           	    {
           	        $T \leftarrow$ Receive T from the parent agent in the BFS-Tree\;
           	    }
           	    Send $T$ to all the child agents in BFS-Tree\;
             \For{$s = 1 ... S_{max}$}{
           	    Synchronously start \Round(True)\;
           	    Wait for Modified\_ALS($\cdot$) to terminate\;
           	    \For{$e_k \in E$}
           	    {
           	        $e_k \leftarrow e_k + \frac{bestCost_{S,k}}{S_{max}}$
           	    }
               }
            $Threshold \leftarrow $G-th best of set $E$\; 
            $SelectedSample \leftarrow \{t_k: e_k\le Threshold\}$\;
            Update $\theta$ using $SelectedSample$\;
        }
        Synchronously start \Round(False)\;

\caption{The DPSA Algorithm}
\label{algo:DPSACE}

\end{algorithm}

The DPSA algorithm consists of two main components: the parallel SA component and the learning component. The parallel SA component (Procedure 3), is an extension of the existing DSAN algorithm that simulates $K$ systems in parallel. We also modify the existing ALS framework to make it compatible with parallel simulation. The other significant component of DPSA is the iterative learning component detailed in Algorithm~\ref{algo:DPSACE}. It starts with a large temperature region and iteratively tries to shrink it down to an OTR of a small length ($\epsilon$). To obtain this, at each iteration, agents cooperatively perform actions (i.e. synchronously simulate parallel SA with different constant temperatures), collect feedback (i.e. the cost yields by different simulations) and use the feedback to update the current temperature region toward the goal region. The underlying algorithm that is used in the learning process is based on cross-entropy importance sampling. However, to make DPSA sample efficiently, we also present modifications that significantly reduce the number of iterations and parallel simulations needed.\par  

In DPSA, the Simulate($\cdot$) procedure (Procedure 3) runs SA on $K$ copies of a system in parallel. This procedure is called in two different contexts: during learning to collect feedback (Algorithm~\ref{algo:DPSACE}: line $11$) and after the learning process has ended (Algorithm~\ref{algo:DPSACE}: line $18$). The main difference is that in the first context the procedure runs a short simulation in $K$ different constant temperatures (one fixed temperature for each copy). In the second case, the function runs for significantly longer and all $K$ systems run on the learned OTR with a fixed scheduler (discussed shortly). Also, in the first case, all copies are initialized with the same random value from the domain. This is done because we want to identify the effect of different constant temperatures on the simulation step, and initializing them with different initial states would add more noise to the feedback. In the second case, we initialize with different random states. Note that, to avoid confusion, we use $x_i$ to refer to the actual decision variable and $x_{i,k}$ to refer to $x_i$ on the k-th copy of the system. The parameter $isLearning$ is used to represent the context in which the procedure was called. Depending on its value, variables of all K systems are initialized and the length of the simulation is set (lines $2-7$) ($S_{len}$ is the simulation length during learning). After that, the main simulation starts and runs for $L$ iterations.\par 

\begin{procedure}[t]
  \caption{Simulate(isLearning )}
  \SetKwFunction{Temp}{Scheduler}
  \SetKwFunction{seter}{AssignmentDecisionSoft}
\SetKwFunction{Neighbour}{Select\_Next}
  \DontPrintSemicolon
  \small
\eIf{isLearning is True}{
            set $x_{i,k}$ $\forall k \in \{1,...,K\}$ to same random value\;
            $L \leftarrow S_{len}$\;
        }
        {
            set $x_{i,k}$ $\forall k \in \{1,...,K\}$ to different random value\;
            $L \leftarrow Remaining Iterations$\;
        }
        \For{$l = 1 ... L$}{
            send value of $x_{i,k}$ $\forall k \in \{1,...,K\}$ to neighbouring agents\;
            receive value of $x_{i,k}$ $\forall k \in \{1,...,K\}$ from neighbouring agents\;
            Update $x_i$ using Modified\_ALS($\cdot$)\;
             \For{$k = 1 ... K$}{
       	        $v \leftarrow $ sample a random value from $D_i$\;
       	        $t_k \leftarrow $\Temp{$l$,$k$,isLearning}\;
       	        $\Delta_k \leftarrow \sum_{n_j\in N_i}^{} Cost_{i,j}(x_{i,k},x_{j,k}) - Cost_{i,j}(v,x_{j,k})$\;
                $P \leftarrow $ min($1,e^{\frac{\Delta_k}{t_k}}$)\;
                set $x_{i,k}$ to $v$ with probability P\; 
           }
        }    
\end{procedure}

At the start of each iteration, each agent $a_i$ shares the current state of each system (i.e. the variable assignment of each system) with their neighbours (Procedure 3: lines $8-9$). Each agent $a_i$ then updates $x_i$ and performs other operations related to Modified\_ALS($\cdot$) (discussed shortly) (line $11$). After that, for each of the K systems, agent $a_i$ picks a value randomly from its domain $D_i$ (line $12$). Afterward, each agent selects the temperature for the current iteration i.e. $t_k$ (line $14$) by calling the function Scheduler($\cdot$). If it is called during the learning context, it is always set to a constant. More precisely, it is set to the k-th value of $T$ (from lines 26 and 29). Otherwise, if the learned OTR is $[T_{min},T_{max}]$; it can be used with a temperature scheduler, for example a linear scheduler (i.e. temperature decreased linearly with time) can be used. To calculate the temperature using a linear scheduler, we can use Equation \ref{lin}.\par
\begin{equation}
    \label{lin}
    T_{min}+(T_{max} - T_{min})\frac{L-l}{L}
\end{equation}

Finally, agents assign the value $v$ to $x_{i,k}$ with the probability $min(1,\exp(\frac{\Delta_k}{t_k}))$ where $\Delta_{k}$ is the local gain (i.e. improvement of the aggregated cost of the local constraints if the assignment is changed) of the k-th system (line $14-15$). If this gain is non-negative, it will always be changed (since $\exp(\frac{\Delta_k}{t_k}) \ge 1$). Otherwise, it will be accepted with a certain probability less than one.\par  
We now describe our extension of ALS. This is used to collect feedback from the simulations during learning and to give DPSA its anytime property. We modify ALS in the following ways:
\begin{itemize}
    \item Since DPSA simulated K systems in parallel, Modified\_ALS($\cdot$) keeps track of the best state and the cost found by each system separately within the duration of a call of the Simulate($\cdot$) procedure. This is used for the feedback. 
    \item Modified\_ALS($\cdot$) also keeps track of the best state and cost across all K systems and all the calls of the Simulate($\cdot$) procedure. Using this, agents assign values to their decision variables. This is used to give DPSA its anytime property. 
\end{itemize}

The first part of the modification can easily be done by running each system at each call with its separate ALS. For the second part, we can have a meta-ALS that utilizes information of the ALSs in the first part to calculate the best state and cost across calls and systems.\par 
We now discuss the learning component (Algorithm~\ref{algo:DPSACE}: lines $3-17$). Here, we start with a probability distribution $\mathcal{G}(\theta)$ over a large temperature region $[T_{min}, T_{max}]$ where $\theta$ is the parameter vector of the distribution. At the start of each iteration, the root agent samples K points (i.e. a set of constant temperatures $T$ from this distribution (Algorithm~\ref{algo:DPSACE}: line $6$)). Agents then propagate this information down the pseudo-tree (Algorithm~\ref{algo:DPSACE}: lines $8-9$). After that, agents synchronously call the Simulate($\cdot$) procedure $S_{max}$ times (Algorithm~\ref{algo:DPSACE}: line $11$). At each call, agents simulate SA in the K sampled constant temperatures (i.e. set T) in parallel. Then using Modified\_ALS($\cdot$), agents collect feedback i.e. cost of the best solution found by the simulation (Algorithm~\ref{algo:DPSACE}: line $12$). Then agents take a mean over $S_{max}$ feedback (Algorithm~\ref{algo:DPSACE}: lines $13-14$). This average should be an unbiased estimation of the expected solution quality i.e. the actual feedback given a large $S_{max}$. Note that in the pseudo-code, we use $bestCost_{S,k}$ to refer to the best cost found by the k-th system in the S-th call. After all the feedback is collected, we use it to update the parameter vector in Algorithm~\ref{algo:DPSACE}: lined $15 - 17$. To this, the first G best sample points are selected (Algorithm~\ref{algo:DPSACE}: lines $15-16$) and then used to update the parameter vector (Algorithm~\ref{algo:DPSACE}: line $19$). In this way, agents iteratively learn the parameter vector $\theta$.\par 
The parameter vector $\theta$ and its update depend on the particular distribution $\mathcal{G}(\cdot)$ used. In this paper, we focus on two particular distributions namely Gaussian $\mathcal{N}(\cdot)$  and uniform $\mathcal{U}(\cdot)$ due to their simplicity. The parameter vector for $\mathcal{N}(\cdot)$ is  $\theta = [\mu, \sigma]$ and consists of the mean and the standard deviation. The new parameter vector is calculated as:\par

\begin{equation}
    \theta_{new} = [\mu(SelectedSample), \sigma(SelectedSample)]
\end{equation}{}
The parameter vector for $\mathcal{U}(\cdot)$ is  $\theta = [T_{min}, T_{max}]$ and consists of the current bound of the temperature region. The new parameter vector is calculated as:\par
\begin{equation}
\small
    \theta_{new} = [min(SelectedSample), max(SelectedSample)]
\end{equation}{}
Finally, $\theta$ is updated as shown in Equation \ref{upd} where $\alpha$ is the learning rate. \par 
\begin{equation}
\label{upd}
    \theta = (1 - \alpha) * \theta + \alpha * \theta_{new}
\end{equation}{}
Updating parameters in the way discussed above requires a considerable number of iterations and samples. To reduce the number of iterations and samples required, we now discuss a few techniques that we used in the experimental section. First, when the number of parallel simulations is small (i.e. the value of K is small), taking a random sample in a large range is not efficient, and taking a stratified sample will cover a large range better. For example, when using $\mathcal{U}(\cdot)$, we may take samples at regular intervals using Equation \ref{uni}:\par
\begin{equation}
    \label{uni}
    T_{min}+(T_{max}-T_{min})*\frac{k-1}{K-1};  k \in \{1,2,...,K\}
\end{equation} 
Second, when $S_{max}$ is small, the estimation of expected cost becomes noisy. To address this, when two sample points produce feedback within a bound of each other, we consider them equally good. We calculate this bound $\gamma$ using Equation~\ref{sen}.\par 
\begin{equation}
\label{sen}
    \gamma = \mathcal{S} * bestCost_{*}
\end{equation}{}
Here, $\mathcal{S}$ stands for sensitivity and is an algorithm parameter and we use $bestCost_*$ to refer to the actual best cost found so far across all the calls of the Simulate(.) procedure. According to this, we may calculate the $Threshold$ in line 19 of Algorithm~\ref{algo:DPSACE} as follows: $Threshold \leftarrow$ G-th best of $E + \gamma$.\par    
Finally, when $R_{max}$ is small, setting the learning rate to a larger value will speed up the learning process. However, if it is set too high, the algorithm might prematurely converge or skip the optimal temperature. Additionally, we can terminate before $R_{max}$, if all the sample points are within $\gamma$ of each other. We now provide an example of the learning process:\par
Suppose, we have $\alpha = 0.4, G = 3, K = 10, \theta = [0.1,100]$ and we use a uniform distribution. In the first round, the sampled points will be (when taken using the regular interval) (Algorithm~\ref{algo:DPSACE}: line 6):
$$
    T = [0.1,11.1,22.2,33.3,44.4,55.5,66.6,77.7,88.8,100]
$$
Let the feedback from each point be (using the modified ALS as described Algorithm~\ref{algo:DPSACE}: line 12):\par
$$
    E = [50,40,30,25,32,42,57,70,95,130]
$$
The selected sample points (top $G = 3$ points) will be (Algorithm~\ref{algo:DPSACE}: lines 15-16):\par
$$
    SelectedSample = [22.2,33.3,44.4]
$$
Finally, the parameter update will be (min and max of $SelectedSample$) (Algorithm~\ref{algo:DPSACE}: lines 17):\par
$$
    \theta_{new} = [22.2,44.4]
$$
$$
    \theta = 0.6*[0.1,100]+0.4*[22.2,44.4] = [8.9,77.8]
$$
This process will repeat until the termination conditions are met.\par 

After the learning process ends, agents call the Simulate($\cdot$) for the final time at line $18$ of Algorithm~\ref{algo:DPSACE}. At this time, the simulation usually runs for longer on the learned optimal temperature region. This concludes our discussion on the learning component.\par

\subsection{The Greedy Baseline Algorithm}
\begin{figure}[!ht]
\centering
  \includegraphics[scale = 1]{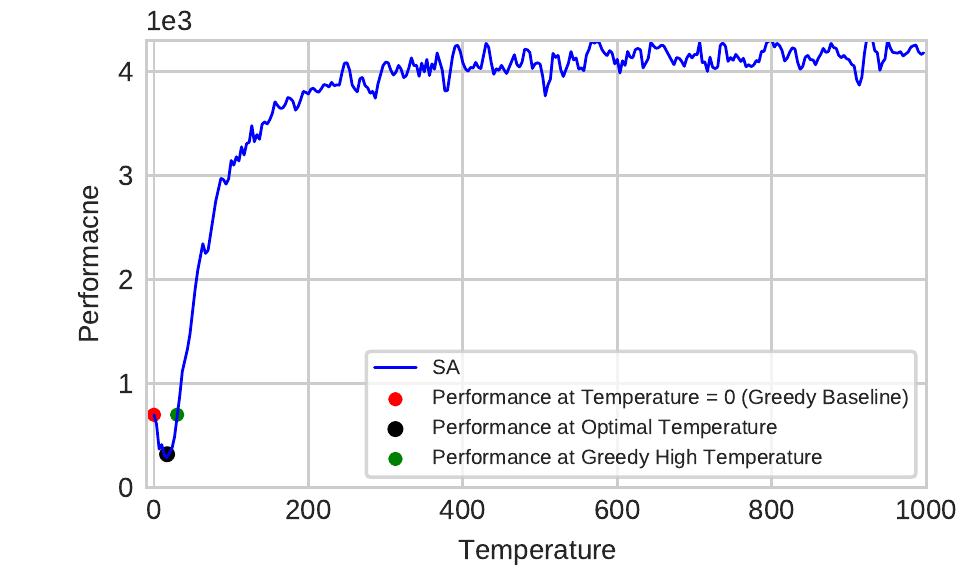}
  \vspace{-3mm}
  \caption{Performance of SA at different constant temperature on Weighted Graph colouring Problem.}
  \label{tmax}
  \vspace{-2.5mm}
\end{figure}

\noindent Much of DPSA's performance depends on the initial parameter vector $\theta$. Nevertheless, it might be the case that prior knowledge about a good initial value of $\theta$ is not available. As discussed earlier, SA behaves randomly\footnote{By random, we mean there is no difference between selecting an assignment from the domain of a variable randomly and applying SA to select the assignment.} when the value of the temperature is high. Now, since DPSA relies on the performance of SA at a different temperature to find an optimal temperature region, such random behaviour significantly affects the performance of the learning component of DPSA. As can be seen in Figure~\ref{tmax}, the average performance of DPSA at the different constant temperatures on the Weighted Graph colouring Problem (see Section 6 for more details). At a high temperature (i.e. above 300), the performance of SA becomes random\footnote{yielding similar average cost as uniform randomly sampled complete assignments.}. Now, imagine a scenario where the range of initial $\theta$ is very large (most of it consists of such high temperatures). In that case, most, if not all, of the sample points of CE might fall in the region where SA behaves randomly. Since this will not create any useful feedback, DPSA would fail to learn a good value of $\theta$. As we discussed in the previous section, DPSA may take stratified sample points. As a consequence, at least one sample is likely to be in the region where the performance of SA is not random (since the first sample point will be near zero). However, even with this, DPSA takes a large number of iterations (R\_max) to find a good temperature region. To address this shortcoming, we introduce a Greedy Baseline (GB) approach that exploits the nature of the temperature vs performance curve of SA to prune away temperature region where SA behaves randomly.\par

To understand how GB works, we again make use of Figure~\ref{tmax}. Here, the Red marker shows the performance of SA at zero temperature. The Black marker shows the performance of SA at the optimal temperature. Finally, the green marker shows the performance of SA at the highest temperature $> 0$ where performance is at least as good as performance at zero temperature. We now formally define this as: 

\textbf{Definition 1.} \textit{A Greedy Baseline (GB) given simulation length $L$} is the expected performance that SA yields at constant zero temperature.\par   
\textbf{Definition 2:} \textit{A Greedy High Temperature ($GHT$)} is the highest temperature $> 0$  where the expected performance of SA is at least as good as GB.\par 

\begin{algorithm}[t]
\DontPrintSemicolon
Initialize variables: $l_{min}, l_{max}, T_{min}, T_{max}$\;
$E \leftarrow \{e_1 = 0,e_2 = 0,...,e_K = 0\}$\;
Synchronously start \Round(True)\;
Wait for Modified\_ALS($\cdot$) to terminate\;
$B \leftarrow E$\;   	    
\While{Conditions not met}{
    $E \leftarrow \{e_1 = 0,e_2 = 0,...,e_K = 0\}$\;
    \eIf{the agent is the root}
       {
           set $T_{min},T_{max}$ to $10^{(l_{min}+l_{max})/2}$\;
       }
       {
           Receive $T_{min},T_{max}$ from the parent agent in the BFS-Tree\;
       }
    Send $T_{min},T_{max}$ to all the child agents in BFS-Tree\;
    Synchronously start \Round(True)\;
    Wait for Modified\_ALS($\cdot$) to terminate\;
    \eIf{$E$ statistically worse than $B$}
       {
           $l_{max} \leftarrow (l_{min}+l_{max})/2$\;
       }
       {
            $l_{min} \leftarrow (l_{min}+l_{max})/2$\;
       }
}
$\theta \leftarrow [\epsilon, 10^{(l_{min}+l_{max})/2}]$
\caption{The Greedy Baseline Algorithm}
\label{algo:DPSAgb}
\end{algorithm}

As discussed earlier, DPSA tries to approximate the OT, while GB tries to approximate $GHT$ using binary search and statistical hypothesis testing. The intuition behind GB is that most trivial/baseline performance yielded by SA is at zero temperature where its behaviour is completely greedy\footnote{i.e. an agent only changes its current assignment if it improves local cost.}. Any temperature where the performance of SA is worse than this is best avoided. By comparing performance at different temperatures with greedy performance, we can prune away such temperature regions. After that, if the remaining temperature region is sufficiently small then DPSA stops the learning phase; otherwise, it can be further narrowed down using the CE method. Finding $GHT$ in this way does not require much prior knowledge (i.e. a good initial value of parameter vector $\theta$). In fact, as we report in Section 6, GB can find $GHT$ starting with an initial parameter vector $\theta$ in the range $[10^{-18}, 10^{18}]$. As this is the largest value handled by a 64-bit integer, we can consider it to be arbitrary for most practical purposes.\par
GB starts by initializing variables (Algorithm~\ref{algo:DPSAgb}: line 1). Here, $l_{min}$ and $l_{max}$ are the log values of the initial temperature region. For example, if we start with an initial temperature region $[10^{-18}, 10^{18}]$, it is set to -18 and +18, respectively. $T_{min}$ and $T_{max}$ is the minimum and the maximum value of the temperature region respectively, where the simulation is run.\footnote{When the simulation is run with a constant temperature, they are both set to the same value.} They are both initialized to 0.\par
Following the initialization, GB finds the Greedy Baseline $B$ by simulating all K systems at zero temperate\footnote{To clarify at zero temperature, when $\Delta_k \ge zero$ the P from Procedure 3 line 15 is considered 1 and when  $\Delta_k < zero$ then P is considered 0.} (Algorithm~\ref{algo:DPSAgb}: Lines 2-5). Recall, E is the best ALS performance of each system. After this, GB iteratively performs a binary search to find GHT. To do this, at each iteration GB first runs all K simulations at the midpoint of the current temperature region (Algorithm~\ref{algo:DPSAgb}: lines 7-13) and collects their ALS performance (Algorithm~\ref{algo:DPSAgb}: line 14). After that, it performs a statistical test to see whether E is statistically worse than the baseline $B$. If this is the case, the midpoint is above GHT and hence $l_{max}$ is set to this midpoint (Algorithm~\ref{algo:DPSAgb}: line 16). Otherwise, it is below or on GH; therefore, $l_{min}$ is set to the midpoint (Algorithm~\ref{algo:DPSAgb}: line 18). There are several candidate statistical tests available. For example in our experiments, we simply checked weather upper bound of the 99\% confidence interval of the average of $B$ is bigger than the lower bound of the 99\% confidence interval of the average of $E$. It is easy to see that with the increase of K, GB will find a better approximation of GHT.\par 
Finally, GB sets the parameter vector $\theta$ to [$\epsilon$, GHT] (Algorithm~\ref{algo:DPSAgb}: line 19). It is worth mentioning that GB should be called between line 2 and 3 of Algorithm~\ref{algo:DPSACE} (DPSA). If this region is sufficiently small, DPSA sets $R_{max}$ to zero (i.e. skipping the learning stage). Otherwise, DPSA applies CE to further narrowing down this region. Since GB prunes away the temperature region where SA does not provide any good feedback, CE can perform efficiently in the pruned region.\par 

\subsection{Complexity Analysis}
\noindent In terms of complexity, the main cost is yielded by the Simulate($\cdot$) procedure. Each iteration of this requires calculating the local gain $\Delta_k$ for $K$ systems. The calculation of local gain requires $O(|N|)$ complexity where $|N|$ is the number of the neighbours. Hence, the computational complexity is $O(K|N|)$ (per iteration and agent). To compare, DSA requires $O(|D||N|)$ (per iteration and agent) where $|D|$ is the size of the domain. Thus, when K = $|D|$, both DSA and DPSA have the same asymptotic complexity. In terms of communication complexity, $K$ variable assignments are transferred at each iteration which gives it $O(K)$ complexity. Finally, agents have to save $K$ local variable assignments, each of which requires $O(|N|)$ memory, meaning the total memory requirement is $O(K|N|)$. It is worth mentioning that the memory requirement of Modified\_ALS($\cdot$) is $O(K|H|)$ where $|H|$ is the height of the BFS\:tree. In Modified\_ALS($\cdot$), while the number of messages remains the same as ALS; the size of each message increases by a factor of\:$K$.   

\section{Empirical Results }
\noindent We now compare the anytime performance of our proposed AED, DPSA\_CE (DPSA only using CE) and DPSA\_GB (DPSA only using GB) algorithms against a number of the most prominent incomplete DCOP algorithms. In particular, we benchmark against DSA-C \cite{fitzpatrick2003distributed, ArshadDistributedSA}, DSAN \cite{ArshadDistributedSA}, GDBA \cite{okamoto2016distributed} and DSA\_SDP \cite{zivan2014explorative}. We selected this set since, as highlighted in Section 2, GDBA and DSA\_SDP are representative of the state-of-the-art local search algorithms, while DSA-C and DSAN serve as a good baseline performance of solution quality of incomplete algorithms. Further, we run these local search algorithms in parallel using the modified ALS in a similar manner as DPSA. This improves solution quality and acts as a fairer comparison. Moreover, the only other population-based DCOP algorithm ACO\_DCOP is used in our benchmark suite. Finally, LSGA\_DSA is chosen as it is the only available hybrid population-based (local search and genetic operation) DCOP algorithm. We selected LSGA\_DSA rather than other variants because it has been reported to produce the best performance \cite{chen2020genetic}. 

We use P = 0.8 for DSA-C because this value of P yields the best performance in our settings (with a few exceptions that are mentioned later). To evaluate GDBA, we use the (N, NM, T) variant, and for DSA\_SDP we consider pA = 0.6, pB = 0.15, pC = 0.4, pD = 0.8. We run 10 instances of these algorithms in parallel in a similar manner as DPSA using the modified ALS. To evaluate ACO\_DCOP, we use the same values of the parameters recommended in \cite{Chen2018AnAA} with one exception for the population parameter of ACO\_DCOP $K$. We vary the value of $K$ from 5 to 20, and pick the one that performs the best within the time limit since ACO\_DCOP suffers from scalability issues. In case of AED, we keep $\alpha = 3$ and the population parameter $ER=1$ and pick the best $\beta$ from integers in $[1,12]$. We will mention specific K for ACO\_DCOP and $\beta$ for AED in their corresponding benchmark sections. For LSGA\_DSA, DPSA\_CE and DPSA\_GB, we set the population size to 10 (similar to the parallel local search execution to make it fair). We use the same parameters as recommended in \cite{chen2020genetic} for LSGA\_DSA. We use $R_{max} = 12, S_{max}=1, S_{lan} =100, \alpha =0.5, \mathcal{S} = .01$ and $\mathcal{G} =3$ with an initial temperature region $\theta = [10^{-3},10^3]$ in case of DPSA\_CE. On the other hand, DPSA\_GB does not require much parameter tuning and we use $S_{len} = 100$ with an initial temperature region  $[10^{-18},10^{18}]$. We use an Intel Core i5-7500 CPU @ 3.40GHz CPU with 8GB RAM to run all the experiments. We cross-checked our results with the results shown in the corresponding papers and code available in the public repositories from the authors and well-known libraries\footnote{\textit{https://github.com/czy920} and \textit{https://frodo-ai.tech}.}.\par 
In the subsequent five subsections, we present five different benchmarks.  All differences shown in the Figures of these subsections are statistically significant for $p-value<0.01$. To numerically compare these algorithms, we will refer to their relative solution cost (RS)\footnote{we use this since calculating the actual optimal cost is infeasible.} which we calculate as:
$$
Relative\ solution\ cost\ of\ Algorithm_i, (RS_i) = \frac{Lowest\ Solution\ Cost}{Solution\ Cost\ of\ Algorithm_i} *100
$$. 
\subsection{Random DCOPs}
\begin{figure}
\centering
\begin{subfigure}[b]{0.7\textwidth}
   \includegraphics[width=1\linewidth]{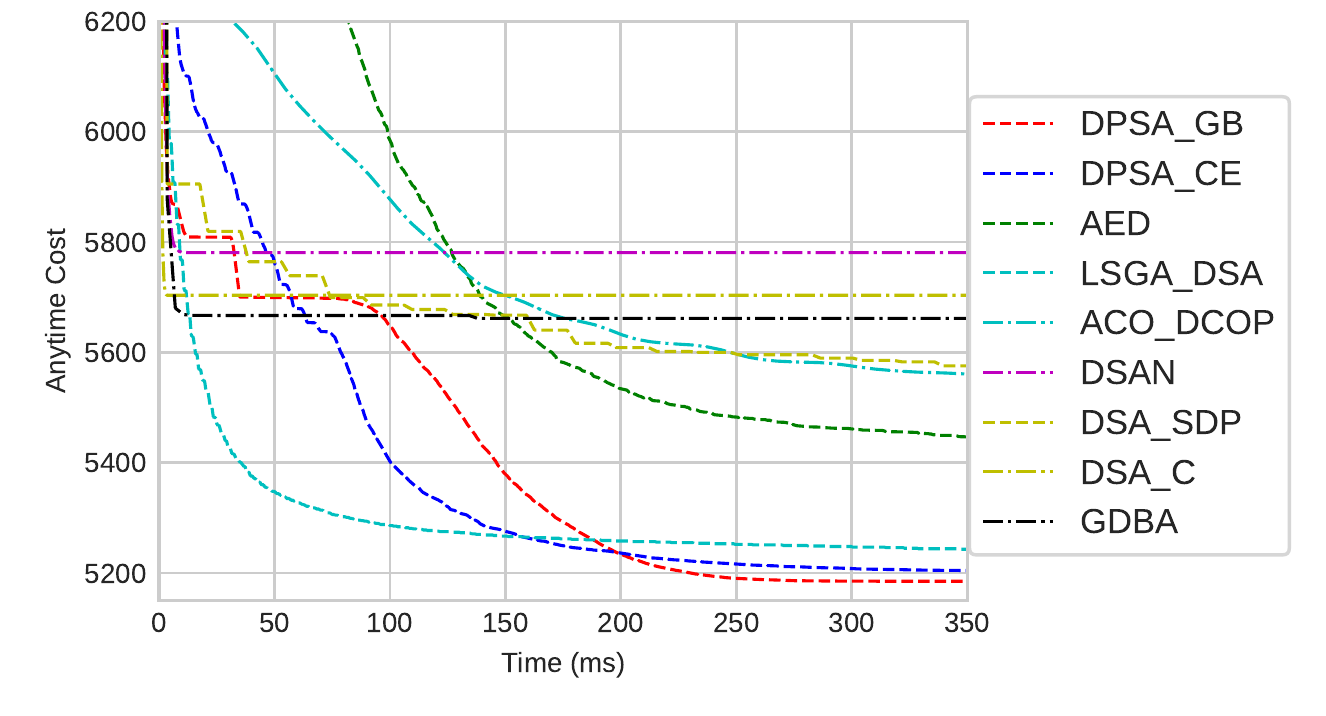}
   \caption{Number of Agents = 70, Network Density = 0.1, Domain Size = 10}
   \label{fig:RD10} 
\end{subfigure}

\begin{subfigure}[b]{0.7\textwidth}
   \includegraphics[width=1\linewidth]{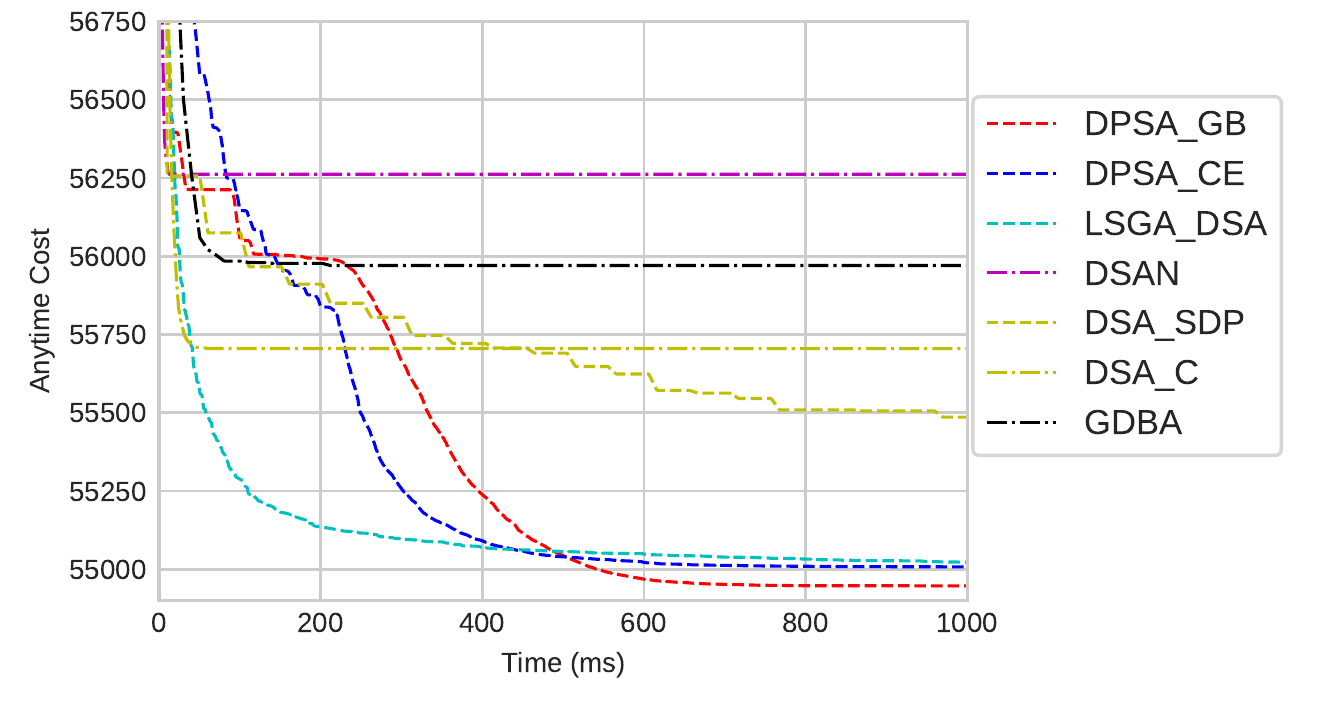}
   \caption{Number of Agents = 70, Network Density = 0.6, Domain Size = 10}
   \label{fig:RD60}
\end{subfigure}
\begin{subfigure}[b]{0.5\textwidth}
   \includegraphics[width=1\linewidth]{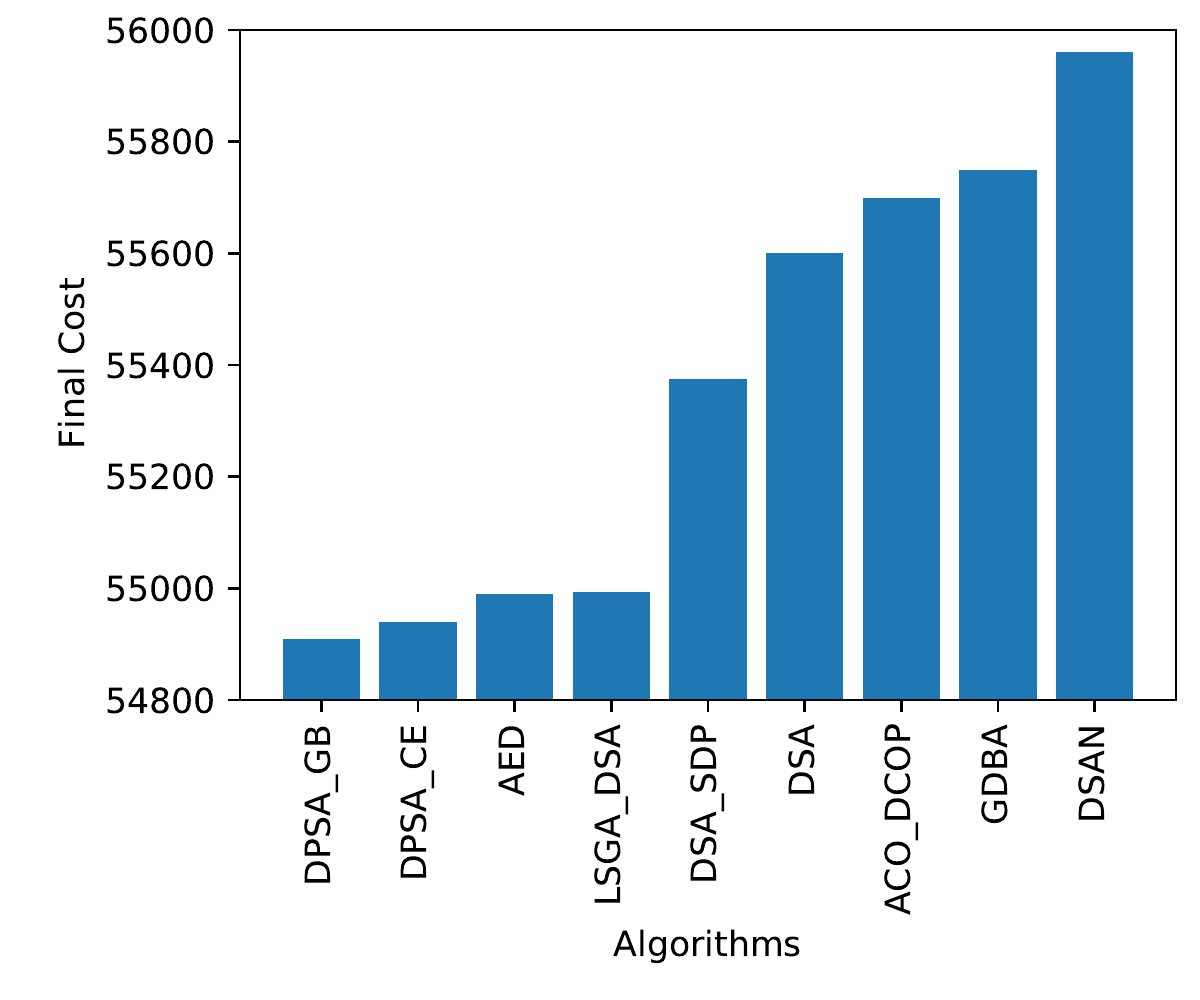}
   \caption{Number of Agents = 70, Network Density = 0.6, Domain Size = 10, 7500 ms}
   \label{fig:Box60}
\end{subfigure}
\caption{Performance on the Random DCOPs Benchmark.}

\end{figure}

 \noindent Random DCOPs \cite{zivan2012max} is one of the most widely used DCOP benchmark problems. To generate problems for this benchmark, we first construct a constraint network using Erd{\H{o}}s-R{\'e}nyi topology \cite{erdHos1960evolution} with 70 agents. We vary the constraint density from 0.1 to 0.6. The cost for each pair of assignments is selected randomly from the interval $[1,100]$. We generate 30 instances of Random DCOPs using this procedure. We then run each algorithm on each instance for 30 times and average the result. The results are used to create the plots that are shown in Figure~\ref{fig:RD10} and Figure~\ref{fig:RD60}.\par 

In the sparse settings, the best result is achieved by DPSA\_GB (RS=100). The two closest algorithms are DPSA\_CE with RS = 99.5 and LSGA\_DSA with RS = 99.1. Here, we use $\beta = 7$ for AED. Even though the above 3 algorithms came within 1\% of their final results after 350 ms, AED was still improving. After the allotted 350 ms, the final RS of AED was 94.9. Even though this is a significant improvement over parallel local search and ACO\_DCOP; AED fell short due to time constraints against the hybrid population-based algorithms. For ACO\_DCOP, we used K=13 which allowed ACO\_DCOP to produce RS = 92.7. Among the local search algorithms, DSA-SDP performed the best with RS = 92.7 and DSAN performed the worst with RS = 89.5.\par   

Even in the dense setting, DPSA\_GB produced the best results (Figure~\ref{fig:RD60}). However, we see LSGA\_DSA and DPSA\_CE produce a similar result with RS = 99.8. The slight variation we observe between DPSA\_CE and DPSA\_GB is because of the annealing effect. Since DPSA\_GB works on a slightly larger temperature region than DPSA\_CE, the higher temperature sometimes helps DPSA\_GB to avoid bad local optima which is not the case with DPSA\_CE. However, this comes at the cost of initial slower optimization. Among other algorithms, we see DSA\_SDP produces the best results ends up yielding RS = 98.9. In both of these settings, DPSA has a clear advantage over all the other algorithms.\par

In Figure~\ref{fig:RD60}, we excluded AED and ACO\_DCOP because within this time they were unable to produce good results. To see how ACO\_DCOP and AED perform in dense settings, we perform the experiment again for the 7500 ms. To make it fair, we restart each of the local search algorithms 500ms apart. For ACO\_DCOP we set K = 20. The result is summarized in Figure~\ref{fig:Box60}. We can see given sufficient time, AED can outperform most of the state-of-the-art algorithms and yields a similar result to LSGA\_DSA. However, DPSA\_GB still produces the best result.\par 

\subsection{Sensor Network Problems}
\begin{figure}
\centering
\begin{subfigure}[b]{1\textwidth}
   \includegraphics[width=1\linewidth]{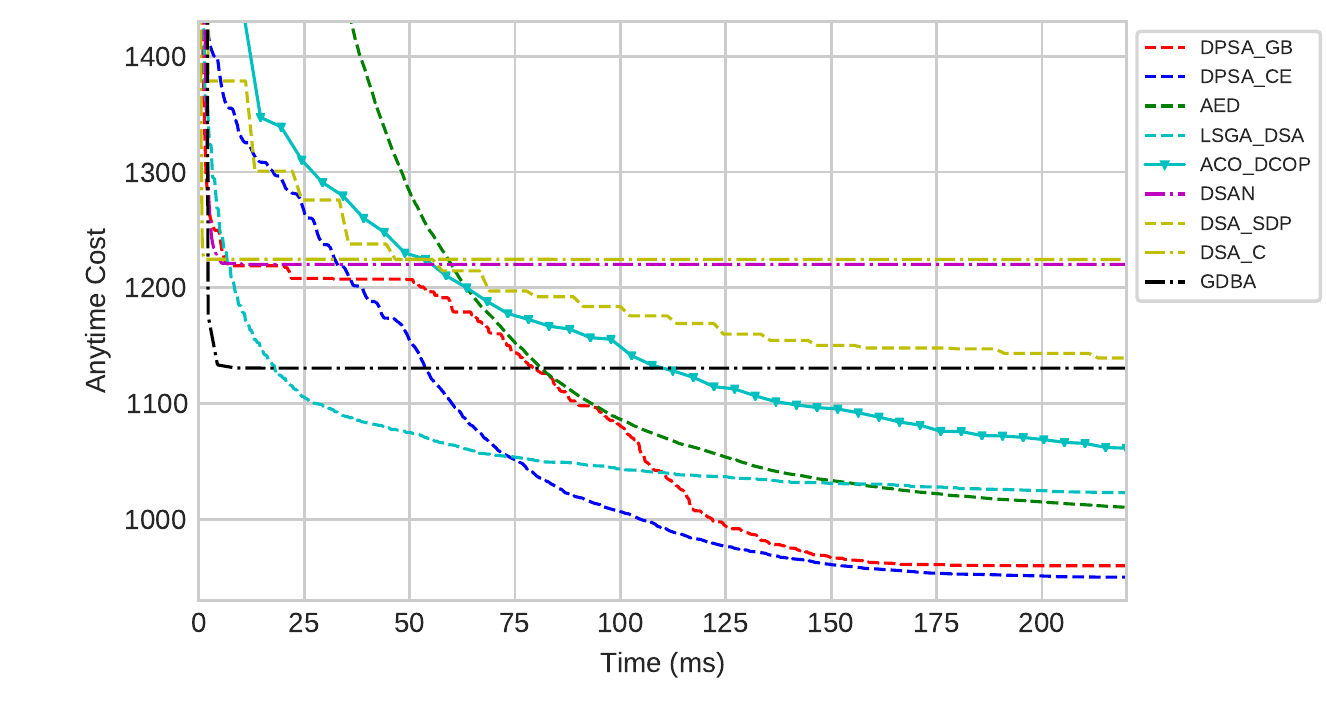}
   \caption{Number of Agents = 49, Domain Size = 12}
   \label{fig:RF49} 
\end{subfigure}

\begin{subfigure}[b]{1\textwidth}
   \includegraphics[width=1\linewidth]{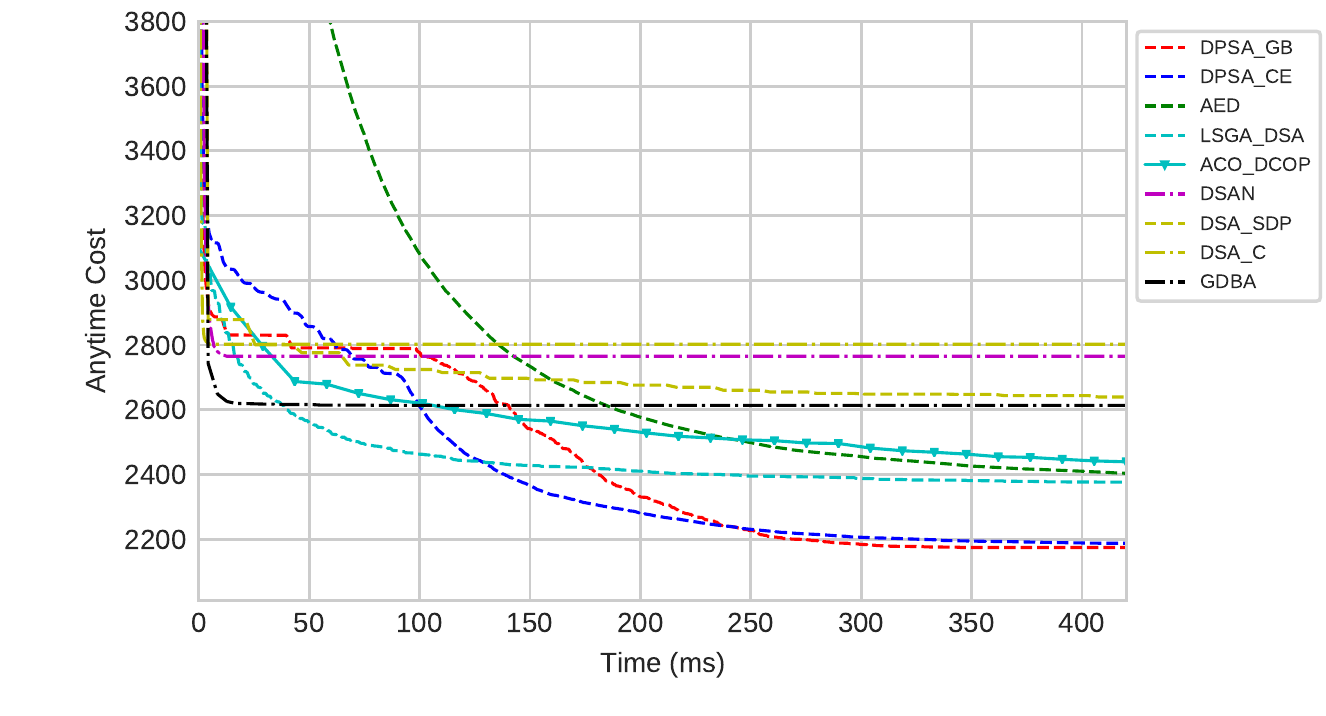}
   \caption{Number of Agents = 100, Domain Size = 12}
   \label{fig:RF100}
\end{subfigure}

\caption{Performance on the Sensor Network Problems.}

\end{figure}

\noindent The sensor network coordination problem, as discussed in \cite{realworld,dgibbs}, involves first responders in an emergency situation deploying mobile sensor agents to create a meshed communication network. The goal is to maximize the signal strength between neighbouring mobile agents in this network. To generate a problem, we arrange the agents in 2D grid cells. Each agent can move and locate themselves in 12 different locations within their cells. This corresponds to their domain. Each agent is constrained with the agents of neighbouring cells. We chose the constraint utilities uniformly from the range [1, 100]\footnote{It can be interpreted as 1 is the lowest signal interference and 100 is highest. The goal is to minimize signal interference.} at random. We generate 30 problems instances with two different grid sizes: $7X7$ (number of agents= 49) and $10X10$ (number of agents= 100). The results are shown in Figure~\ref{fig:RF49} and Figure~\ref{fig:RF100}.\par  
In the small grid, we see all three of DPSA\_CE, DPSA\_GB (RS = 98.9) and AED (RS = 94.0) outperform the state-of-the-art, while DPSA\_CE produces the best results. In this setting, scalability is not an issue for AED due to the low constraint density. This results in AED outperforming LSGA\_DSA (RS= 93.1) within the 250 ms time limit. On the other hand, for ACO\_DCOP, we set K = 10 and within 250 ms, it yields RS = 89.6. Among the other algorithms, GDBA produce the best results (RS=84.0) while DSA-C produce the worst (RS = 77.6).\par

In the larger setting, DPSA\_GB produces the best result while DPSA\_CE is closest (RS = 99.0). In this setting, scalability starts to become an issue for AED (RS = 90.0) and yields a slightly worse result than LSGA\_DSA (RS = 90.4), while ACO\_DCOP produces RS = 88.9. However, given more time, both ACO\_DCOP and AED outperform LSGA\_DSA. In this experiment, we see in low-density settings, AED does not suffer from scalability issues and so produces results that outperform all of the competing algorithms.\par  

\subsection{Scale-Free Network Problems}
\noindent We now consider the scale-free network problem. This is a similar to Random DCOPs, except we use the Barabási-Albert network topology model \cite{Barab} to generate our constraint networks. To that end, we start with a random tree of 20 agents ($m_0$), we then connect each new agent with a set of size $m$ randomly selected existing agents. For the sparse problems, we set $m$ to 3, and for the dense problems, we set $m$ to 12. We use 100 agents to create our network, the remaining parameters are the same as used on the random DCOPs setting. \par

\begin{figure}
\centering
\begin{subfigure}[b]{1\textwidth}
   \includegraphics[width=1\linewidth]{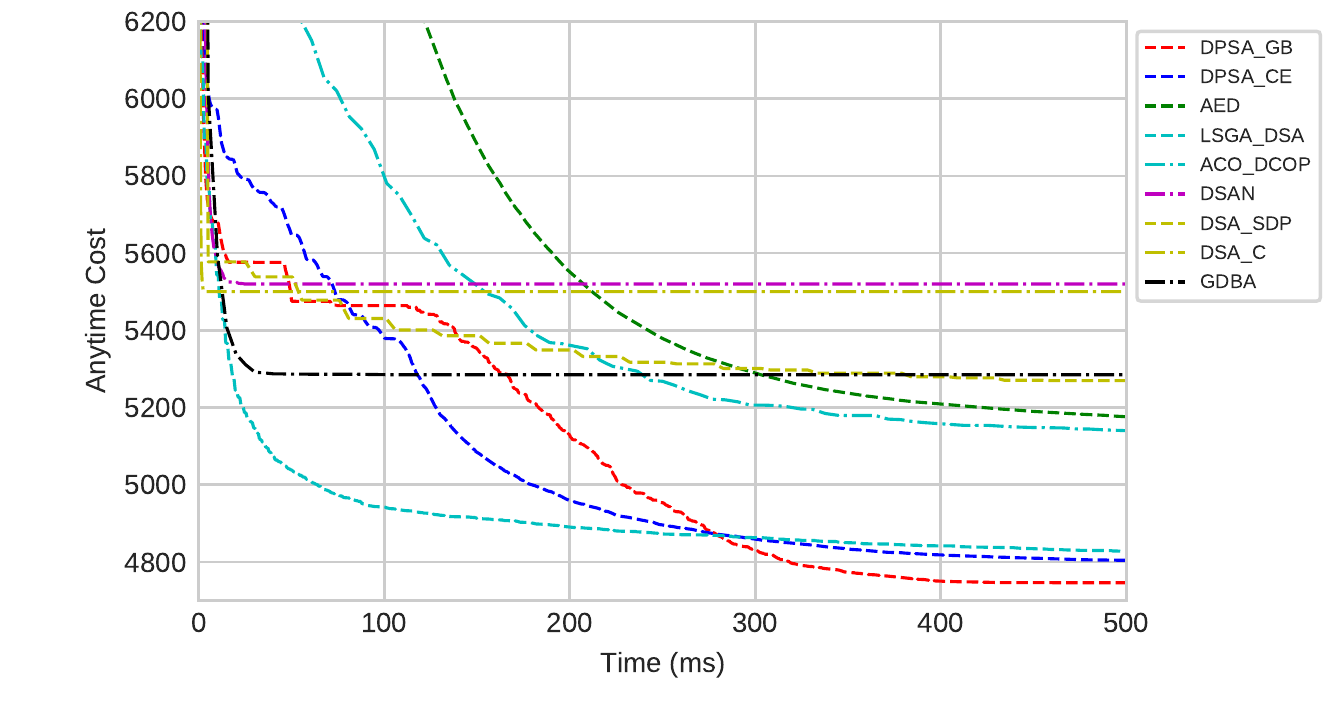}
   \caption{Number of Agents = 100, Domain Size = 10, $m_0 =20$, $m = 3$}
   \label{fig:SF3} 
\end{subfigure}

\begin{subfigure}[b]{1\textwidth}
   \includegraphics[width=1\linewidth]{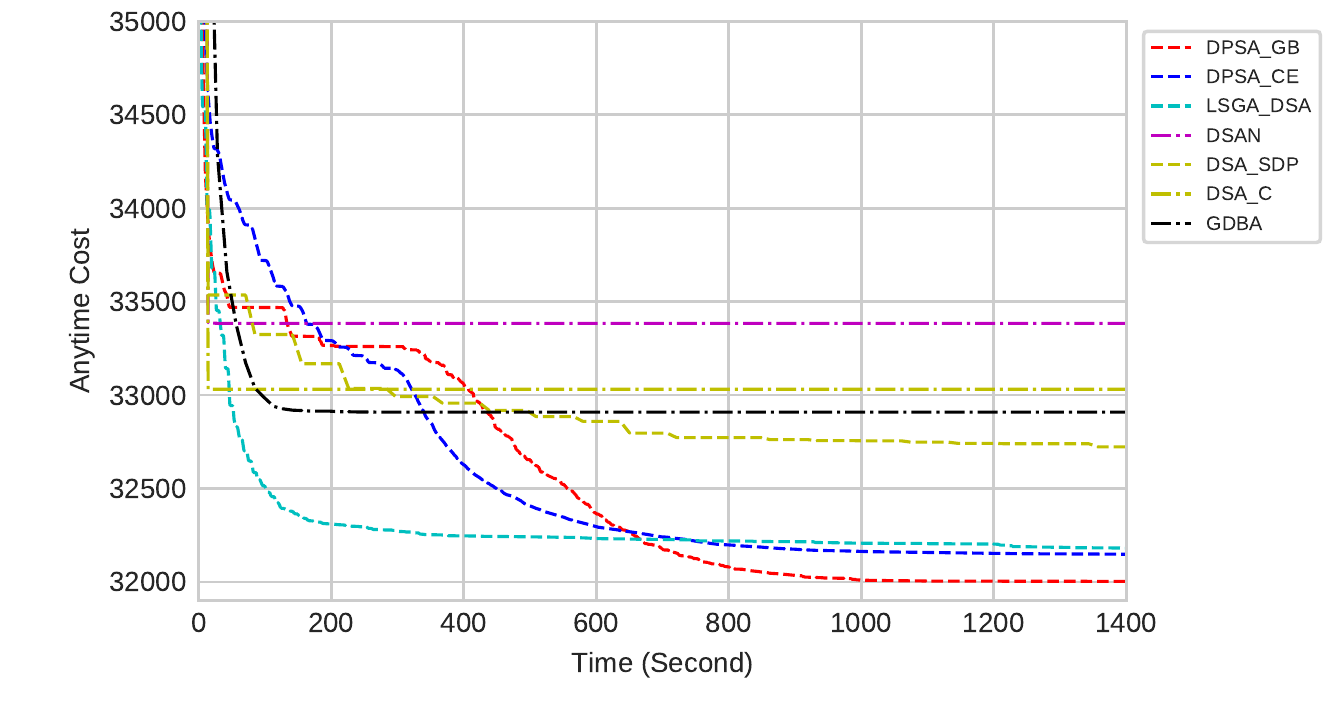}
   \caption{Number of Agents = 100, Domain Size = 10, $m_0 =20$, $m = 12$)}
   \label{fig:SF12}
\end{subfigure}

\caption{Performance on the Scale-Free Network Problems.}

\end{figure}
We see a similar performance difference between LSGA\_DSA (RS = 98.2 on m=3 and RS = 99.5 on m =12), DPSA\_CE(RS = 98.8 on m=3 and RS = 99.5 on m=12) and DPSA\_GB (RS = 100 on m=3 and RS = 100 on m=12) in both settings. Although better than local search algorithms, AED (RS = 91.7) is able to outperform other population-based algorithms including the ACO\_DCOP(K = 13) (RS = 92.0). Among other algorithms DSA\_SDP produces the best results (RS = 91.2 on m=3 and RS = 97.7 on m=12) and DSAN yields the worst performance (RS = 85.6 on m=3 and RS = 95.8 on m=12). Overall, DPSA\_GB produces significantly better results than the other competing algorithms including DPSA\_CE.
A key insight we can draw from this experiment along with previous two experiments is that different widely used network topologies (random, grid and scale-free) do not have any adverse impact on the solution quality produced by DPSA. Neither does density (as we varied in random DCOP and scale-free network settings) or size (as we varied in sensor network setting) of the constraint network under these typologies have any negative effects. In all cases, DPSA outperformed the state-of-the-art.\par 

\subsection{Weighted Graph colouring Problems}
\noindent The weighted graph colouring problem with 3 colours is another widely used benchmark problem for DCOPs. To generate a problem, we first construct a constraint network using Erd{\H{o}}s-R{\'e}nyi topology \cite{erdHos1960evolution} with 120 agents. We vary the constraint density from 0.05 to 0.2. The penalty for choosing the same colour with a neighbouring agent of the constraint graph is selected randomly from $[1,100]$.\par
In this problem, DPSA (CE and GB) shows overwhelmingly excellent results. In the sparse setting, the closest competitor is LSGA\_DSA (RS = 66.1). Even though AED (RS = 49.5) outperforms all the local search algorithms and the population-based algorithm, it could not outperform the hybrid population-based algorithms. Among other algorithms, GDBA does the best (with RS = 45.2) and DSA-C performs the worst (with RS = 31.0).\par
In the dense setting, we also see a similar trend though the differences in the results are smaller. DPSA\_CE still manages to produce the best performance. However, in this setting, LSGA\_DSA's result is close yielding RS = 99.2. Among the other local search algorithms, GDBA does the best (with RS = 97.0) and DSAN performs the worst (with RS = 94.3). For DSA, we set P = 0.7 because for P = 0.8 it does not produce a good result, especially in the dense settings. In this case, neither AED nor ACO\_DCOP can produce a good result and hence they are omitted from the dense setting.\par
Overall, DPSA shows its excellent performance by outperforming all the algorithms with a large margin of $34\%-66\%$. This highlights how profoundly temperature affects the performance of SA and getting it correct has a significant impact on solution quality.\par 

\begin{figure}
\centering
\begin{subfigure}[b]{0.7\textwidth}
   \includegraphics[width=1\linewidth]{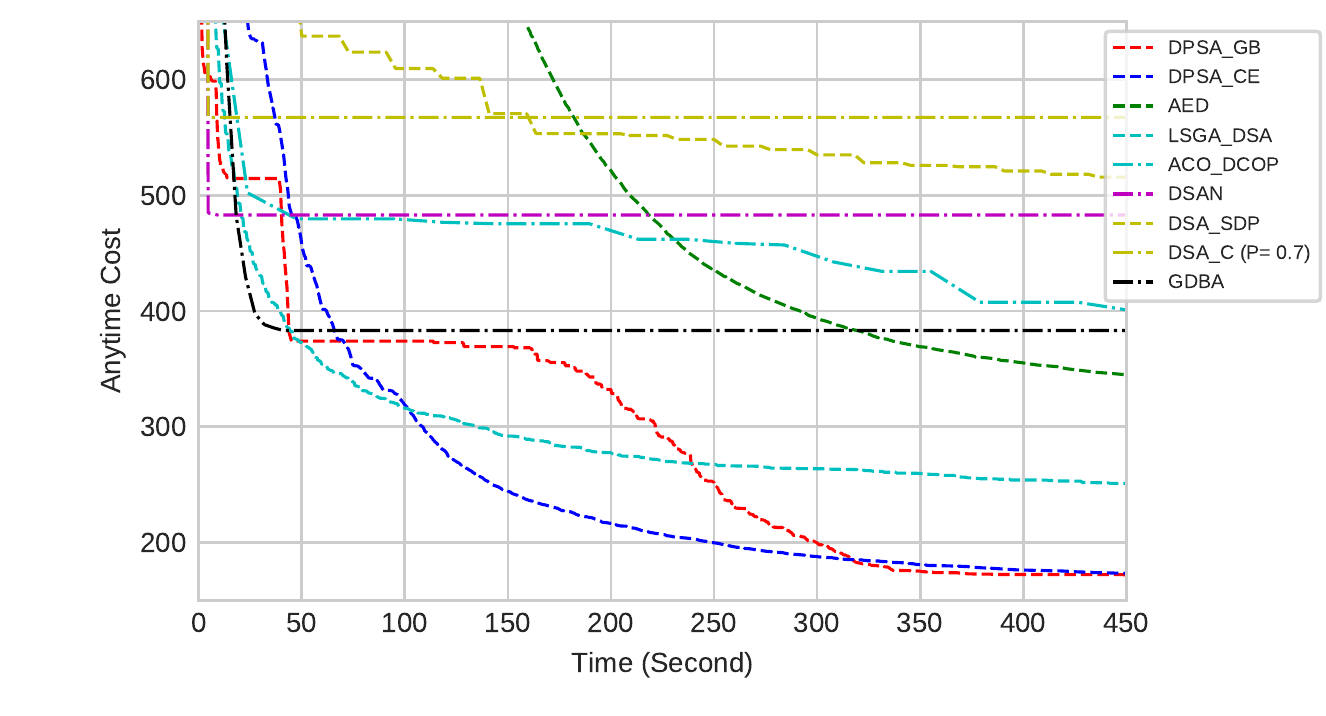}
   \caption{Number of Agents = 120, colours = 3, Network Density = 0.05}
   \label{fig:WGC05} 
\end{subfigure}

\begin{subfigure}[b]{0.7\textwidth}
   \includegraphics[width=1\linewidth]{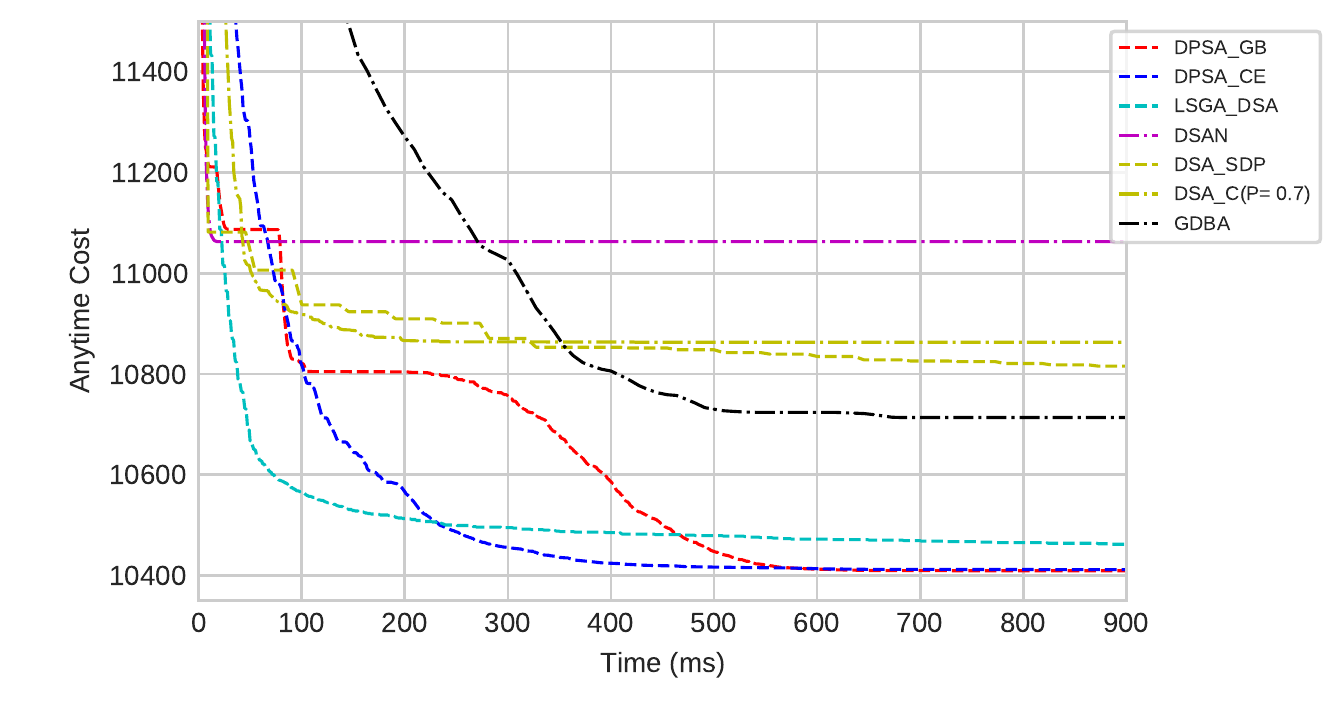}
   \caption{Number of Agents = 120, colours = 3, Network Density = 0.20}
   \label{fig:WGC20}
\end{subfigure}

\begin{subfigure}[b]{0.5\textwidth}
   \includegraphics[width=1\linewidth]{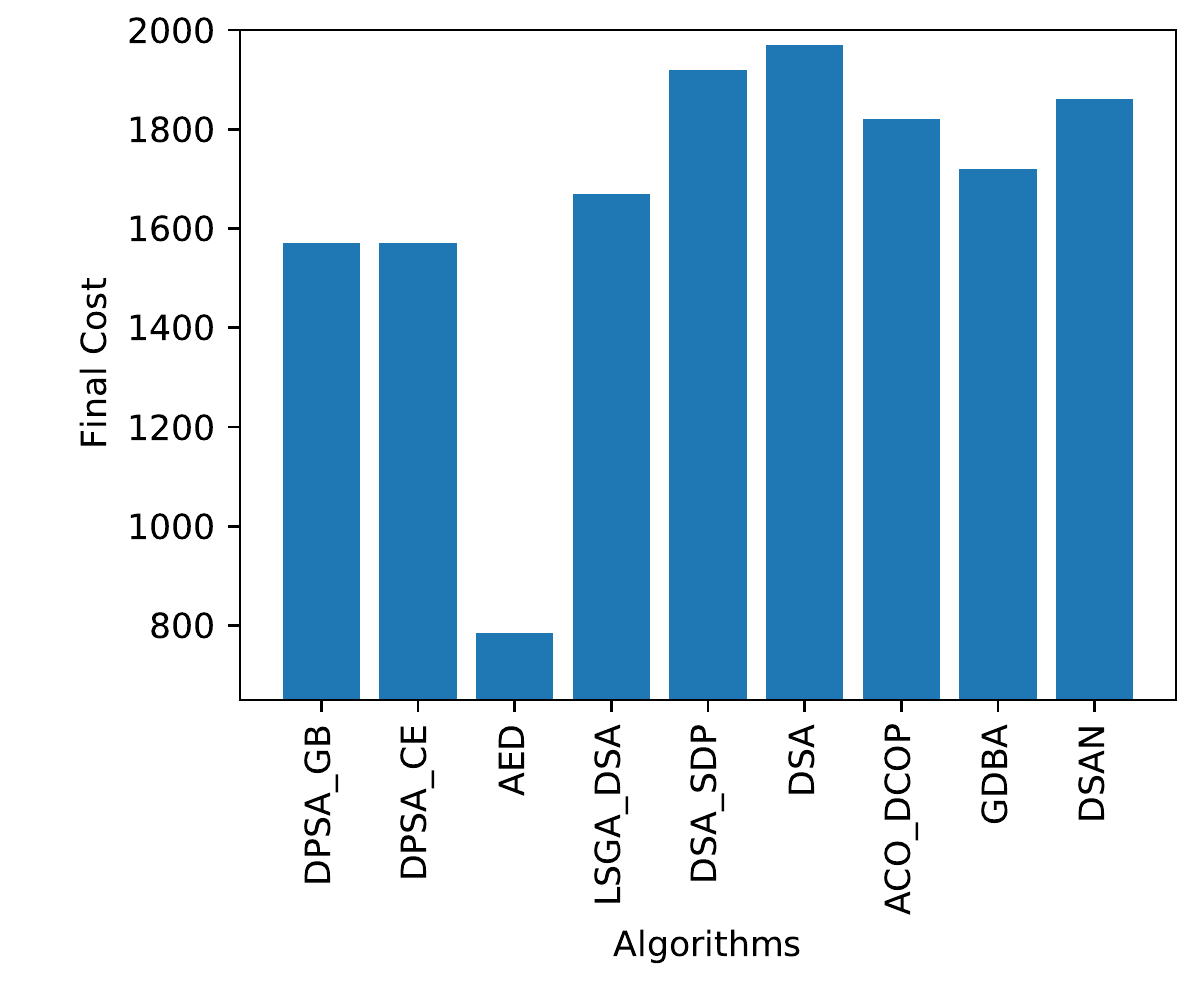}
   \caption{Number of Agents = 120, colours = 3, Network Density = 0.05, With Global Constraints}
   \label{fig:GC05}
\end{subfigure}

\caption{Performance on the Weighted Graph colouring Problems.}

\end{figure}
We now evaluate the performance of competing algorithms by imposing an additional global constraint on top of the weighted graph colouring problems used to depict the results of Figures 6a and 6b. In this experiment, we consider an additional constraint that each colour can be assigned to at most 40 agents on the generated problem of 120 agents having a network density of 0.05. For each colour, each additional assignment above 40 would incur an additional cost of 500 to the global cost. In other words, if each colour is assigned to exactly 40 agents or less, no additional cost will incur. This can be interpreted as a soft variant of the well-known All Diff global constraints \cite{van2001alldifferent}. We keep all the configuration similar except we set $\alpha = 8$ and $\beta = 2$ for AED. The result is shown in Figure~\ref{fig:GC05}.\par
AED outperforms all the competing algorithms by a large margin (i.e. RS = 50.1 to 39.8, up to 75\% improvement in solution quality compared to the competing algorithms). This is expected as other algorithms, that includes both versions of DPSA, are not able to handle the global constraint. This is because in those algorithms, agents do not have information about the complete assignments and therefore they are unaware whether they are breaking the global constraint or not. On the other hand, AED has access to the entire assignment and can enforce such global constraints. It is worth noting that we also run this benchmark by modifying competing algorithms and letting all the agents broadcast their assignments. This scenario is analogous to running the benchmark on a complete graph. This results in a significant increase in computation and communication costs  (more preciously, it causes a similar effect if the density is increased by $1/0.05 = 20$ times). Also, within the given amount of time (i.e. 400 ms), all the algorithms perform even worse. This demonstrates that AED can be successfully applied to scenarios where there exists global constraint(s).\par     

\subsection{Target Tracking Problems}
\noindent Target tracking is an important and widely studied problem for surveillance, monitoring applications and data collection. Some instances that have been studied in the literature include collecting data using autonomous underwater robotic vehicles \cite{under}, monitoring points of interest in an unknown environment \cite{dcopmst}, wide-area surveillance \cite{farinelli2014agent} and collecting data for security resource allocation. In this problem, we have a set of sensor agents (e.g. UAVs, robots) tracking a set of targets in order to collect accurate information about them. For tracking targets, agents can take several different actions. For example, if they are mobile agents, they can move and locate themselves in such a way that minimizes their uncertainty about the target or improve the quality of data gathered.\par 

\begin{figure}
\centering
\begin{subfigure}[b]{1\textwidth}
   \includegraphics[width=1\linewidth]{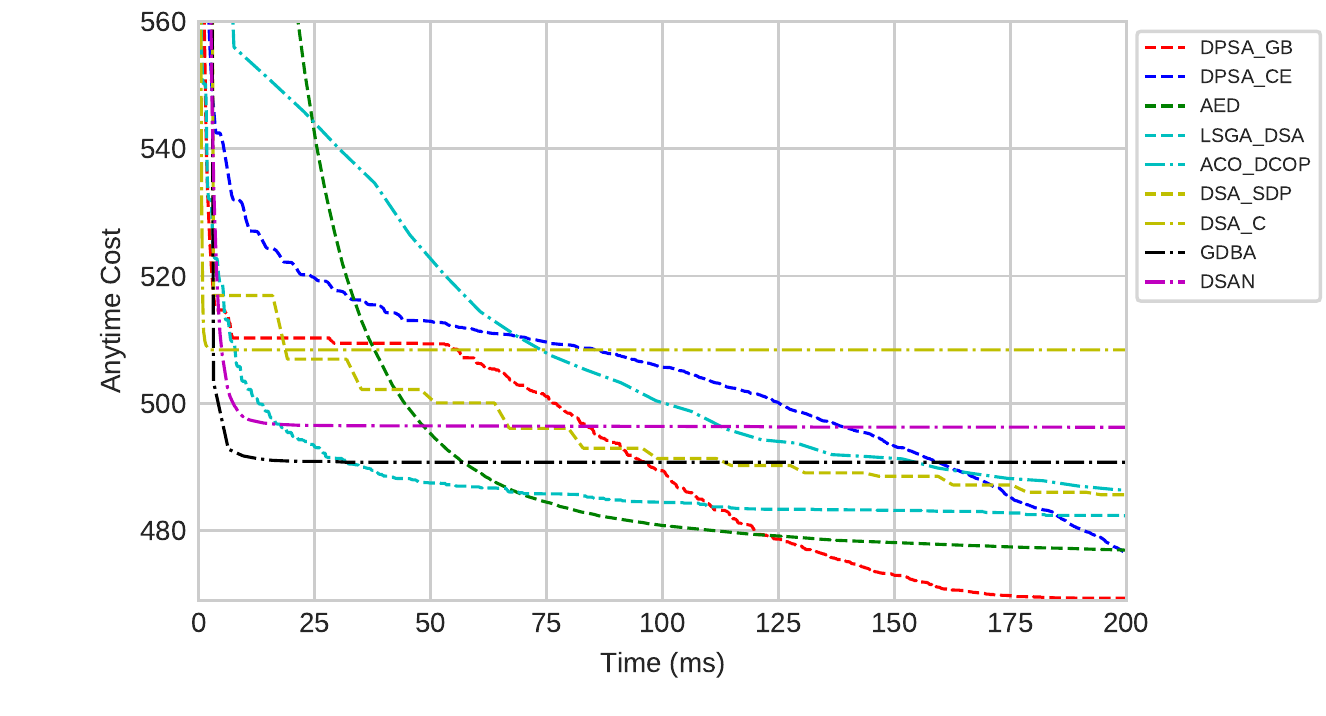}
   \caption{Number of Agents = 49, Number of Positions = 25}
   \label{fig:UAV49} 
\end{subfigure}

\begin{subfigure}[b]{1\textwidth}
   \includegraphics[width=1\linewidth]{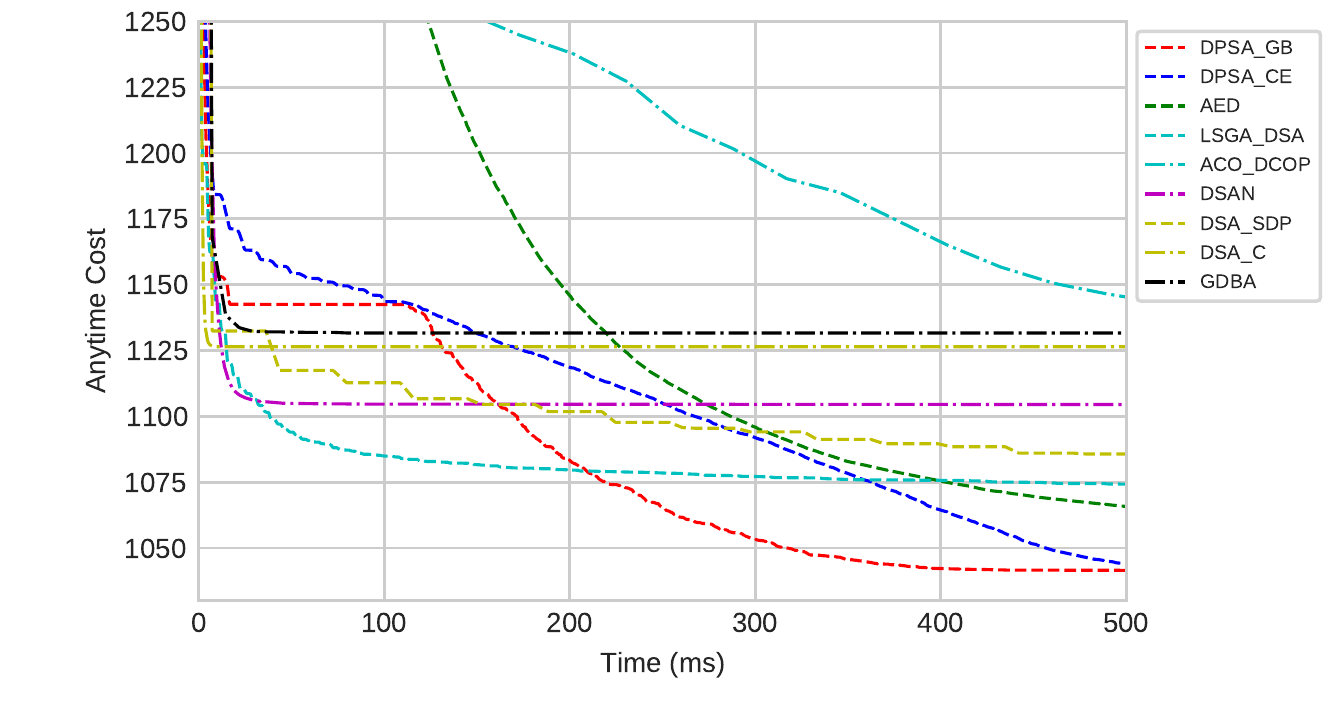}
   \caption{Number of Agents = 10, Number of Position = 25}
   \label{fig:UAV100}
\end{subfigure}

\caption{Anytime performance on the Static Target Tracking Problems.}

\end{figure}

Notably, several ways have been studied to formalize this scenario using the DCOP framework \cite{dcopmst, farinelli2014agent}. Here, we take a simplified approach to generate problems. Each sensing agent is mobile and is located in a grid cell. Within a grid cell, each agent has 25 different positions (by dividing the cell into 25 squares) where it can move to. Hence, in this problem, each variable $x_i$ denotes the position of agent $a_i$ within its grid cell. There is a constraint between each pair of neighbouring agents which represents a target that the agents need to collect data from. For each target, we select one of 10 random positions on the border of those two agents' grid cell randomly. Each target $t_{i,j}$ between agent $a_i$ and $a_j$ also has an importance rating $IR_{i,j}$ selected uniform randomly from $[1,30]$. Each pair of agents in position ($x_i,x_j$) associated with each target yields a different amount of data loss $DL_{i,j}(x_i,x_j)$, which is a number between $(0,3]$ based on average distance from the target rated $(0,2]$ and random environmental factor rated $(0,1]$. The constraint cost yields by each assignment pair ($x_i,x_j$) is calculated as $IR_{i,j}*DL_{i,j}(x_i,x_j)$. Agents cooperatively try to minimize the total data loss. We generate 30 problems instances 
with two different grid sizes: $7X7$ (number of agents= 49) and $10X10$ (number of agents= 100). The results are shown in Figure~\ref{fig:UAV49} and Figure~\ref{fig:UAV100}.\par  
 
In the smaller grid, we see all three of DPSA\_GB, DPSA\_CE\footnote{$G = 2$ for this benchmark} (RS = 98.7) and AED (RS = 98.7) outperform the state-of-the-art, while DPSA\_GB produces the best results. However, given more population, AED ($\beta = 11$) can match the results of DPSA\_GB. In this setting, we do see a large random temperature region affecting the performance of the DPSA\_CE method and as such, it takes a significant amount of time to find a good solution compared to AED and DPSA\_GB. After R\_max times out, DPSA\_CE needs to iterate through a large temperature region linearly. Notably, DPSA\_GB avoids this by pruning the bad temperature region using greedy baseline. The nearest competitor is LSGA\_DSA which converges after yielding RS = 97.3. Among the local search algorithms, DSA-SDP does the best (RS = 97.7) and DSA\_C performs the worst (RS = 92.5).\par
In the larger setting, DPSA\_GB produces the best result while DPSA\_CE is the nearest (RS = 99.2). Even though AED (RS = 98.2) outperforms the previous state-of-the-art; it is not able to outperform DPSA with ER =1. However, the increasing value of ER significantly decreases the performance gap. The best among the other algorithms is LSGA\_DSA(RS = 97.4) doing slightly worse than AED. Among the other algorithms, we see a similar trend.\par 
\subsection{Discussion}
\noindent In this section, we have covered a wide variety of DCOPs settings to evaluate our algorithms against the state-of-the-art. These settings cover most commonly used network topologies (i.e. random, grid and scale-free), different numbers of agents (from 49 up to 120), a wide range of constraint densities (from 0.036 up to 0.600) and different cost structures. In all of these settings, DPSA markedly outperforms the state-of-the-art. On the other hand, AED outperforms the previous state-of-the-art in smaller settings (i.e. lower numbers of agent and/or low density) even yielding similar performance to DPSA (Sections 6.2 and 6.5). In the larger settings, AED outperforms state-of-the-art local search algorithms and population-based algorithms but could not outperform the hybrid population-based algorithm LSGA\_DSA. In some cases, this is due to time constraints and given more time AED does outperform LSGA\_DSA (Section 6.1). However, we also show a scenario where AED can be more efficient than hybrid population-based algorithms such as LSGA\_DSA or DPSA. Specifically, in problems where there are global constraints such as All Different constraint, AED is significantly more efficient since it maintains complete assignments (individuals) and can directly enforce such constraints that the other competing algorithms and DPSA cannot.\par

To compare our proposed algorithms, in smaller settings both AED and DPSA yields similar performance. However, in large settings, DPSA was more scalable than AED  which is a result of the local search like communication structure of DPSA. On the other hand, settings where there are global constraints, AED is more efficient than DPSA. Between DPSA\_CE and DPSA\_GB, we can see in most cases the final solution cost yielded by these algorithms are similar. However, DPSA\_CE has a good initial anytime cost which is achieved by exploiting prior knowledge. This can be useful for example in dynamic settings where problem instance from a one-time step to others does not significantly vary and prior knowledge can be efficiently transferred. On the other hand, DPSA\_GB does not require any prior knowledge and/or much parameter configuration. This is especially useful when it is not possible to directly tune parameters. For example, consider the experiment of Section 6.2 in which first responders in an emergency situation deploying mobile sensor agents.\par 

To tie this all together, these experiments show population-based algorithms do have advantages in terms of solution quality over other classes of incomplete algorithms. This advantage is not simply the effect of keeping multiple candidate solutions in parallel (since we run all the incomplete algorithms using modified ALS) rather effectively exploiting the population (e.g. learning good parameters or keeping most promising solutions) plays a significant role.\par   

\section{Conclusions and Future Work}
\noindent We have focused on population-based approaches for DCOPs and introduced two new algorithms. The first one, AED, is based on evolutionary optimization. AED mutates individuals (i.e. complete assignments) of the population utilizing local information and selects which individual should survive (for further mutation) based on global information. We also propose a hybrid population-based algorithm DPSA that has high scalability compared to existing population-based algorithms. DPSA runs several SAs in a distributed manner in parallel and uses global information from different runs to come up with a good temperature to operate the algorithm on. For calculating a good temperature region, we propose two methods. The first one requires good prior knowledge to succeed (i.e. CE) and the second one does not (i.e. GB). The former has the advantage that it has good initial anytime performance and can exploit prior knowledge when available and the latter has the benefit that it does not require much parameter tuning or prior knowledge. We have shown through extensive experimental evaluation that DPSA outperforms the state-of-the-art in a wide variety of settings. On the other hand, AED outperforms the state-of-the-art local search algorithms and the existing population-based algorithm namely ACO\_DCOP in all our benchmarks while outperforming existing hybrid population-based algorithm (i.e. LSGA) in some benchmarks. We also show that AED can be applied to a scenario where there are global constraints significantly more efficiently than other competing algorithms. Most importantly, our work demonstrates that evolutionary optimization can be effective for DCOPs and hybridization with local search algorithms can be a solution to the scalability issue of population-based DCOPs solver.\par 

In future, we want to explore population-based sampling algorithms. This is especially interesting because existing population-based algorithms do not provide any quality guarantee and population-based sampling algorithms have the potential to provide such guarantees while also producing high-quality approximate solution within a practical amount of time. Finally, there are many variants of SA and algorithms inspired by SA that has been proposed over the last four decades to improve speed or solution quality of SA (e.g. Parallel Tempering \cite{Swendsen1986ReplicaMC}, Stochastic Tunneling \cite{Wenzel1999StochasticTA}, and Threshold Accepting \cite{tacc}). We want to study this class of algorithms for further improvement of SA based DCOP solvers.\par   
\section*{Acknowledgements}
\noindent This paper extends our previously published works on population-based DCOP algorithm \cite{AED,DPSACE}. This research is partially supported by the ICT Innovation Fund of Bangladesh Government.






\bibliographystyle{elsarticle-num-names}
\bibliography{sample.bib}







\end{document}